\documentclass[journal]{IEEEtran}
\usepackage{enumerate}
\usepackage{xspace}
\usepackage[pdftex]{graphicx}
\usepackage{amsmath}
\usepackage{array} 
\usepackage{booktabs} 
\usepackage{multirow}
\usepackage{cite}
\usepackage{url}

\ifCLASSINFOpdf
\else
\fi
\ifCLASSOPTIONcompsoc
  \usepackage[caption=false,font=normalsize,labelfont=sf,textfont=sf]{subfig}
\else
  \usepackage[caption=false,font=footnotesize]{subfig}
\fi

\makeatletter
\DeclareRobustCommand\onedot{\futurelet\@let@token\@onedot}
\def\@onedot{\ifx\@let@token.\else.\null\fi\xspace}
\def\eg{\emph{e.g}\onedot} 
\def\ie{\emph{i.e}\onedot}

\def\etal{\emph{et al}\onedot}
\makeatother

\newsavebox\CBox 
\def\textBF#1{\sbox\CBox{#1}\resizebox{\wd\CBox}{\ht\CBox}{\textbf{#1}}}

\begin{document}
%
\title{Generalized Local Optimality \\for Video Steganalysis in Motion Vector Domain}
%
%
%

\author{Liming~Zhai\textsuperscript{\rm 1,2},
        Lina~Wang\textsuperscript{\rm 1},
        Yanzhen~Ren\textsuperscript{\rm 1},
        and~Yang~Liu\textsuperscript{\rm 2}\\
        \textsuperscript{\rm 1}Wuhan University, China  ~\textsuperscript{\rm 2}Nanyang Technological University, Singapore
}

\maketitle

\begin{abstract}
The local optimality of motion vectors (MVs) is an intrinsic property in video coding, and any modifications to the MVs will inevitably destroy this optimality, making it a sensitive indicator of steganography in the MV domain. Thus the local optimality is commonly used to design steganalytic features, and the estimation for local optimality has become a top priority in video steganalysis. However, the local optimality in existing works is often estimated inaccurately or using an unreasonable assumption, limiting its capability in steganalysis. In this paper, we propose to estimate the local optimality in a more reasonable and comprehensive fashion, and generalize the concept of local optimality in two aspects. First, the local optimality measured in a rate-distortion sense is jointly determined by MV and predicted motion vector (PMV), and the variability of PMV will affect the estimation for local optimality. Hence we generalize the local optimality from a static estimation to a dynamic one. Second, the PMV is a special case of MV, and can also reflect the embedding traces in MVs. So we generalize the local optimality from the MV domain to the PMV domain. Based on the two generalizations of local optimality, we construct new types of steganalytic features and also propose feature symmetrization rules to reduce feature dimension. Extensive experiments performed on three databases demonstrate the effectiveness of the proposed features, which achieve state-of-the-art in both accuracy and robustness in various conditions, including cover source mismatch, video prediction methods, video codecs, and video resolutions.
\end{abstract}


%
\IEEEpeerreviewmaketitle

\section{Introduction} \label{sec:intro}
%
%
%
%
\IEEEPARstart{S}{teganography} and steganalysis are a pair of antagonistic technologies engaged in a cat-and-mouse game fashion,  where the steganography aims to covertly transmit secret messages by embedding them into inconspicuous carriers, and the steganalysis tries to expose the presence of hidden messages in suspicious carriers. What connects the above two opposite parties is message carrier, on which the embedding mechanisms of steganography and the discriminative features of steganalysis rely heavily in practice. Commonly used carriers for steganography and steganalysis are digital media including image, video and audio, among which the digital video has gained more attention recently for following reasons.

First, digital videos are easily accessible carriers in today’s digital media era. Digital video has become ubiquitous and pervasive due to the high popularity of video devices and the wide variety of online video applications, making video stream the main traffic flow in the IP network. According to Cisco Visual Networking Index \cite{cisco2019cisco}, the global IP video traffic will account for 82\% of all IP traffic by 2022, up from 75\% in 2017. Second, digital video has sufficient spatial-temporal information, which provides more embedding space than other types of carriers \cite{Tew2014Overview}. Third, video sharing and broadcast through social networking is conductive to conceal the steganography behavior. It is reported that 400 hours of videos are uploaded to YouTube every minute worldwide \cite{Youtube2021}, acting as a natural camouflage for transmitting videos with hidden messages. These merits of video carrier promote the prosperity of video steganography, and large numbers of video steganographic methods or tools have emerged continuously \cite{Liu2019Overview}, posing a severe challenge to video steganalysis.

The video steganography can be grouped into various categories according to the used coding entities at different compression stages, such as spatial pixels \cite{Cox1997SS, Hartung1998SS}, intra prediction modes \cite{Yang2011PreModes, Nie2018PreModes}, inter partition modes \cite{Kapotas2008MBpar, Yang2012MBpar, Zhang2014MBpar}, motion vectors (MVs) \cite{Xu2006MV, Fang2006MV, Aly2011MV, Cao2011PME, Yao2015MV, Cao2015PMEopt, Zhang2015Preserved, Zhu2018UED}, DCT coefficients \cite{Ma2010Drift, Cao2018CBD}, quantization parameters (QPs) \cite{Shanableh2012MRFMO, Wang2020CEC}, entropy coding syntax \cite{Liao2012CAVLC, Young2008CABAC} and their combinations \cite{zhai2019MDE}. Among these categories, the MV based steganography has attracted great interest due to its large embedding capacity, high security and lossless message extraction \cite{Tew2014Overview}.

The MV based steganography embeds messages into MVs by modifying the values of one and/or two components of MVs. This embedding is performed in the inter coding of video compression, in accompanying with prediction error (PE) adjustment to avoid video quality degradation. The development of MV based steganography can be divided into three stages. The MV based steganography in the first stage is \emph{quality-preserved} methods, which modify the candidate MVs while maintaining the video quality, such as selecting large MV magnitudes \cite{Xu2006MV}, appropriate MV phase angles \cite{Fang2006MV} and small PEs \cite{Aly2011MV}. The second-stage MV based steganography is \emph{statistics-preserved} methods, whose aim is to modify the MVs with minimal impact on video statistics. Typical methods include perturbing motion estimation with suboptimal MVs \cite{Cao2011PME} and altering MVs with slight changes on MV distributions and PEs \cite{Yao2015MV}. To achieve little statistical deviation, some coding schemes, like wet paper codes \cite{Fridrich2005WPC} and syndrome-trellis codes \cite{Filler2011STC}, are also integrated into the embedding. In the third stage, the MV based steganography can be characterized as \emph{local optimality-preserved} methods from the coding point of view. These methods first assess the preservation of MV optimality when suffering MV modifications, based on which the embedding cost for each MV is designed and assigned, and finally embed the messages by using syndrome-trellis codes. Popular methods in this type measure the MV optimality using coding costs of  MVs, including the sum of absolute difference (SAD) \cite{Cao2015PMEopt} and Lagrangian cost \cite{Zhang2015Preserved}. Recently, the combination of video statistics and local optimality is also employed for MV based steganography \cite{Zhu2018UED}.

The MV based steganography has gone through a process which moves the focus from the coding-independent video entities to the intrinsic properties of video coding. Similarly, the video steganalysis in the MV domain also has such a developmental trend. Steganalytic features play a crucial role in steganalysis. Early video steganalytic features are designed drawing on the ideas of image steganalysis, such as characteristic function moments of MV residue distributions \cite{Su2011CF, Deng2013SecondOrderCF}, first-order or high order co-occurrence matrices using MV residues \cite{Wu2014Joint, Tasdemir2016STRM}, and calibration based features \cite{Deng2012LPKRM, Cao2012MVRB, Wang2015IMVRB}. These steganalytic features performs well against the first-stage MV based steganography, but are often incompetent for the later ones which consider more about steganographic security. Besides, they also have some constraints in feature extraction, such as continuous and non-subdivided inter macroblocks, as discussed in \cite{Zhai2017CCF, Zhai2019MVC}. To better reveal the embedding traces in MVs, the coding optimality associated with MVs are explored to construct steganalytic features. The representative method is AoSO \cite{Wang2014AoSO}, which assumes that the original MV is locally optimal, and uses SAD as a criterion to identify whether or not a MV has been changed. A similar methodology was also present in \cite{Ren2014SPOM}, where the SAD based local optimality together with recompression calibration is used to extract features. However, these local optimality based features become ineffective when this ideology is adopt by an adversary \cite{Cao2015PMEopt, Zhang2015Preserved, Zhu2018UED}. As a fightback against the local optimality-preserved methods, subsequent video steganalysis \cite{Zhang2017NPE} suggests using more accurate coding criterion to check the local optimality of MVs. In \cite{Zhang2017NPE}, the rate-distortion cost instead of SAD is utilized to form steganalytic features, achieving significant detection performance.

The local optimality based features have become a prevailing approach for video steganalysis in MV domain, owing to the fact that any locally optimal MVs undergoing changes will be shift to locally non-optimal ones in a high probability, regardless of embedding mechanism. Therefore, the local optimality based features rely heavily on the estimation of local optimality for MVs. However, the accuracy of estimation for local optimality in existing works is still far from the requirements. The SAD based local optimality \cite{Wang2014AoSO, Ren2014SPOM} only focuses on the distortion cost, but neglects the bit-rate cost associated with MVs\footnote{Distortion cost refers to the pixel difference. Bit-rate cost is the bit length required to encode the motion vector difference. See Section II for details.}. Although the rate-distortion based optimality \cite{Zhang2017NPE} considers both the above costs simultaneously, the bit-rate cost is estimated on an unreasonable assumption. For example, when calculating the motion vector difference between the MV and its predicted motion vector (PMV), the \cite{Zhang2017NPE} assumes that the PMV is unchanged. If a changed PMV is used to estimate the bit-rate cost, inaccurate estimation result may occurs (see the detail examples in Section III-B).

In this paper, we propose to construct steganalytic features for video steganalysis by using generalized local optimality. We adopt the rate-distortion cost to measure the local optimality. To identify the steganographic changes of MVs more accurately and stably, we generalize the normal local optimality in two aspects. First, The bit-rate cost (also the rate-distortion cost) is determined jointly by MV and PMV. The PMV may be changed or remain unchanged after embedding, and the uncertainty of PMV will affect the estimation of local optimality for the corresponding MV. Therefore, we suggest checking the local optimality of MVs under the conditions of various possible changes of PMVs, and generalize the local optimality from a static estimation to a dynamic one. This generalization aims to reduce the interference of the PMV changes on rate-distortion costs. Second, the PMV is a special case of MV, whose changes are also reflected in the PMV, so the change of PMV can also be served as an indicator of steganographic embedding. As a result, we suggest checking the local optimality not only for the MVs but also for the PMVs, and generalize the local optimality from MV domain to PMV domain. This generalization aims to increase the diversity of steganalytic features. Finally, we design two types of steganalytic features based on the two generalizations of local optimality, and also propose symmetrization rules to reduce feature dimension. We perform extensive experiments to evaluate the detection accuracy, robustness and complexity of our method, and all the results prove its advantages over existing steganalytic methods.

The contributions of this paper are summarized as follows:

\begin{enumerate}[ 1)]
\item We discover the underlying issue in normal local optimality, and generalize the concept of local optimality in two aspects, including estimating the local optimality of MVs from the fixed PMVs to the dynamic PMVs, and extending the local optimality from the MV domain to the PMV domain.

\item We construct two types of steganalytic features, including the generalized local optimality features in MV domain and those in PMV domain, both of which eliminate the interactions among neighboring MVs, and thus improve the detection accuracy. To our best knowledge, this is the first study to explore the PMVs for video steganalysis.

\item We propose feature symmetrization rules based on the distributions of generalized local optimality, and the symmetrization rules significantly reduce the feature dimensions without sacrificing detection accuracy.

\item The proposed generalized local optimality increases the diversity of steganalytic features while does not obviously affect the computational complexity, indicating its efficiency for practical applications.
\end{enumerate}

The organization of this paper is as follows. Section II briefly introduces the video inter coding related to our method. Section III elaborates on the two types of generalized local optimality. The subsequent Section IV describes the details of feature construction. In Section V, we present and discuss the experimental results. Finally, Section VI concludes the paper.

\section{Preliminaries}  \label{sec:pre}

Our video steganalytic method can be applied to any video coding standards that adopt rate-distortion optimization for inter coding. In this section, we use H.264/AVC \cite{JVT2003Draft} as an example to introduce the preliminary knowledge of video inter coding related to our feature design. We choose H.264/AVC because it is still the most commonly accepted video coding standard, which holds an 82\% market share of video codecs in 2019 \cite{EncoderReport2018}, making it widely applicable for video steganography \cite{Tew2014Overview, Liu2019Overview}.

The H.264/AVC employs a block-based hybrid video coding framework. For a current block in the inter frame, the motion estimation (ME) searches for a predicted block in the previously coded frame (\ie, reference frame) such that the predicted block is most similar to the current block. The similarity of the two blocks is usually measured by the sum of absolute difference (SAD) or the sum of absolute transformed difference (SATD), which represents the visual distortion introduced by video coding. 

\begin{figure}[t]
  \centering
  \includegraphics[width=0.9\columnwidth]{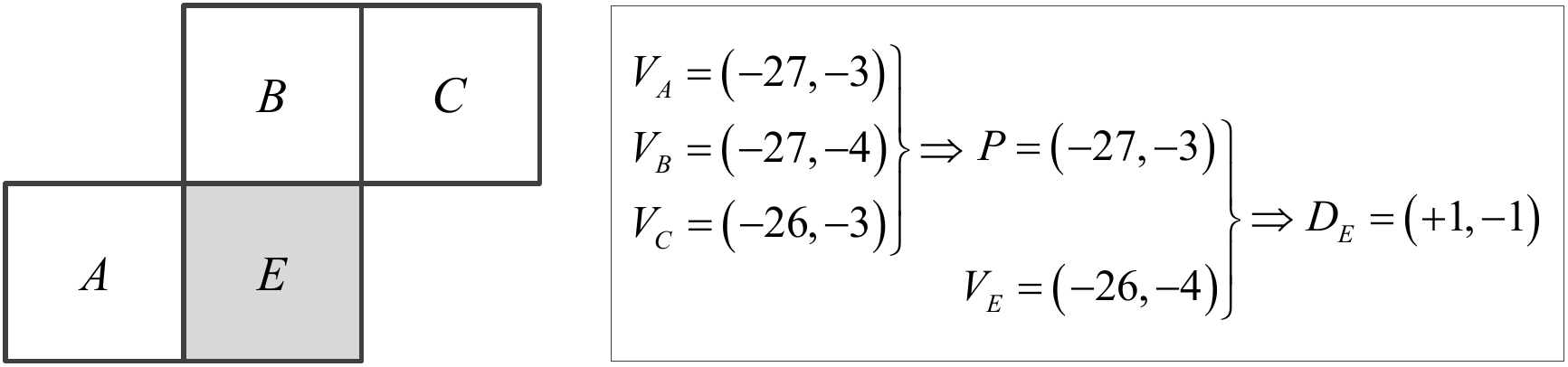}
  \caption{The locations of neighboring blocks (left) and the calculation of predicted motion vector (PMV) and motion vector difference (MVD) (right).}
  \label{fig:Neighbor_PMV}
\end{figure}

The spatial offset between the current block and the predicted block is called motion vector (MV). The MV indicates the position of the predicted block. To reduce bit-rate, the motion vector difference (MVD), which is the difference between the MV and the predicted motion vector (PMV), is encoded into video bit-stream. For the PMV, it is the starting MV for ME, and is usually defined as the median value of the MVs of three neighboring blocks adjacent to the current block, including left, top, and top-right blocks \cite{JVT2003Draft} (see the left figure in Fig. \ref{fig:Neighbor_PMV}).

Visual distortion and bit-rate are the two most crucial criteria for video coding. These two criteria are mutually conflicting, and the process of finding a good trade-off between them is known as rate-distortion optimization (RDO), which is achieved by minimizing the following Lagrangian function
\begin{equation}
  J = {\cal D}\left( {E,{E_V}} \right) + \lambda {\cal R}\left( {V - P} \right),
  \label{eq:rdo}
\end{equation}
where $J$ is the rate-distortion cost, $V$ and $P$ stand for the MV and the PMV, $E$ denotes the current block, ${E_V}$ denotes the predicted block indicated by $V$, ${\cal D}\left( { \cdot , \cdot } \right)$ represents the distortion cost measured by SAD or SATD, ${\cal R}\left(  \cdot  \right)$ represents the bit-rate cost, and $\lambda $ is the Lagrangian multiplier defined as $\lambda  = \sqrt {0.85 \times {2^{{{(Q - 12)} \mathord{\left/
 {\vphantom {{(Q - 12)} 3}} \right.
 \kern-\nulldelimiterspace} 3}}}} $, in which $Q$ denotes the quantization parameter (QP) \cite{Wiegand2003RDO}.

For efficient video coding, practical video codecs often use fast block matching algorithms \cite{Hussain2019Survey} for ME, so the final $J$ has a minimum value in a local search area. This is so-called the local optimality of MVs. Apparently, a changed MV may lose its local optimality after steganographic embedding, which is the foundation of local optimality based features for video steganalysis.

\begin{figure}[t]
  \centering
  \subfloat{\includegraphics[width=0.5\columnwidth]{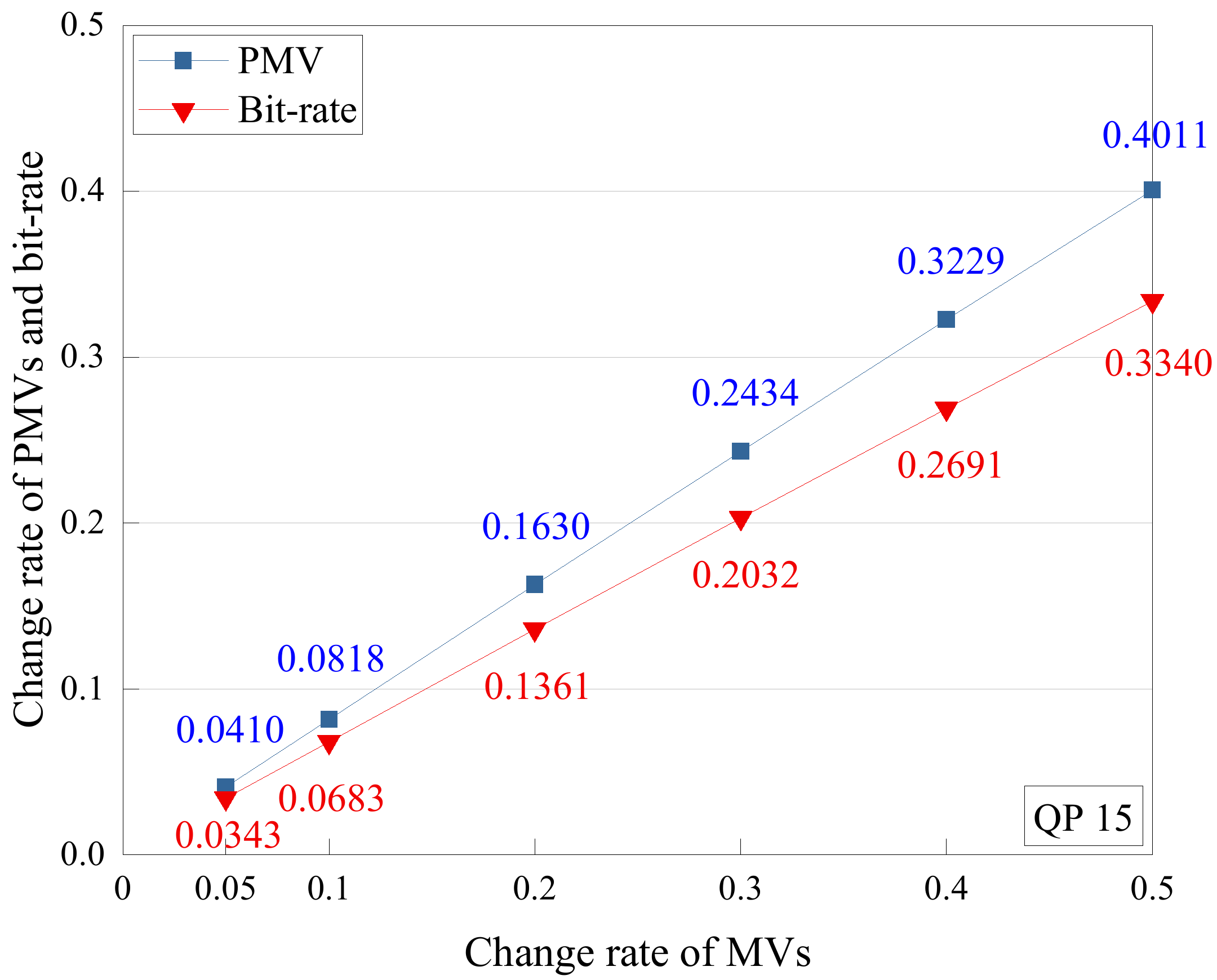}\label{fig:PMV_bitrate_a}}
  \hfil
  \subfloat{\includegraphics[width=0.5\columnwidth]{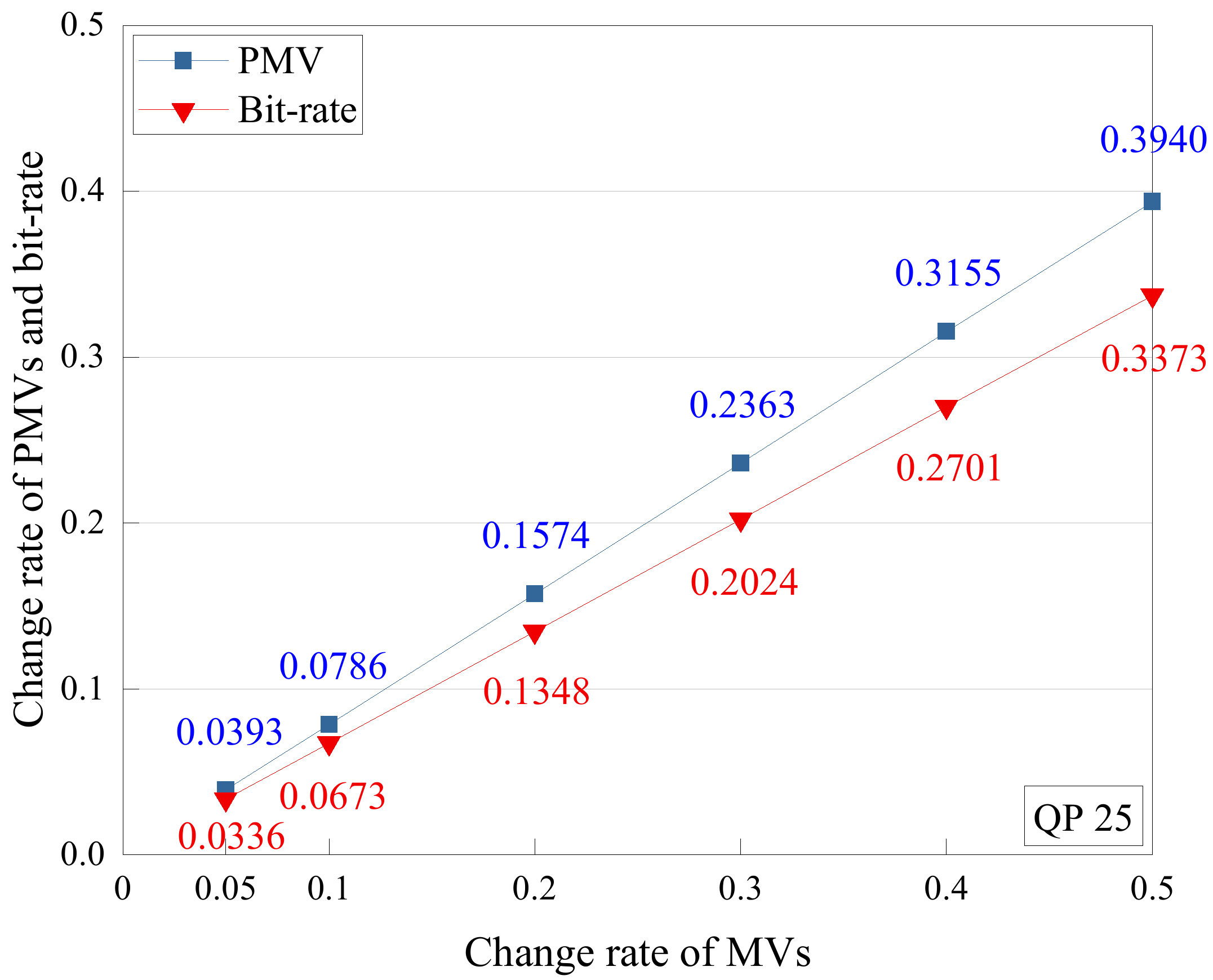}\label{fig:PMV_bitrate_b}}
  \caption{Change rates of PMVs and bit-rates.}
  \label{fig:PMV_bitrate}
\end{figure}

\section{Generalized Local Optimality}  \label{sec:glo}

This section first describes the embedding effect on PMV and bit-rate, and then analyzes the four cases of rate-distortion cost, followed by the definition of generalized local optimality for MV and PMV, and the discussion of the changes of generalized local optimality caused by steganography.

\subsection{Embedding Effect on PMV and Bit-rate}  \label{sec:glo_a}
The RDO is adopted in nearly all video coding standards, so we use rate-distortion cost to measure the local optimality of MVs, as did in \cite{Zhang2017NPE}. Although the rate-distortion cost can capture the changes of MVs more accurately than SAD \cite{Wang2014AoSO, Ren2014SPOM}, it is affected by more factors.

According to (\ref{eq:rdo}), the rate-distortion cost is the weighted sum of distortion and bit-rate, in which the distortion is only controlled by MV, while the bit-rate is jointly determined by MV and PMV. Therefore, when using the rate-distortion cost to check the local optimality of MVs, not only the MVs but also the PMVs need to be considered.

The PMV is a special case of MV, and is formed by the median prediction of three neighboring MVs \cite{JVT2003Draft}. So the changes of neighboring MVs may also lead to the change of the corresponding PMV. To avoid the decoding error, every PMV after steganographic embedding should be updated by the neighboring MVs to calculate the new MVD, which will be encoded into the bit-stream of the stego video.

To describe the relationship between embedding and PMV more intuitively, we use Fig. \ref{fig:Neighbor_PMV} to illustrate the above process. The left figure in Fig. \ref{fig:Neighbor_PMV} shows the locations of current block $E$ and its three neighboring blocks $A$, $B$ and $C$, whose MVs are denoted by ${V_E}$, ${V_A}$, ${V_B}$ and ${V_C}$, respectively. The right figure in Fig. 1 shows the values of four MVs and the calculation results of PMV $P$ and MVD ${D_E}$ corresponding to ${V_E}$. The $P$ is the median of ${V_A}$, ${V_B}$ and ${V_C}$, and if the embedding change ``$+1$'' is only made to the horizontal component of ${V_A}$ or ${V_B}$ ($-27$ is modified as $-26$), the PMV will be changed from $P = \left( { - 27, - 3} \right)$ to $P = \left( { - 26, - 3} \right)$, and the MVD will be changed from ${D_E} = \left( { + 1, - 1} \right)$ to ${D_E} = \left( { 0, - 1} \right)$. For other examples, such as changing the horizontal components and/or vertical components with different combinations of MV numbers and change types, the PMV and the MVD may also change correspondingly.


Meanwhile, the bit-rate may also change with the PMV. The change rates of PMVs and bit-rate under different change rates of MVs are shown in Fig. \ref{fig:PMV_bitrate}. The videos used in this experiment are 36 standard test video sequences randomly selected from DB1 (see Section \ref{sec:exp_setup} for details about this database), and are further H.264/AVC compressed with QP 15 and 25. The stego videos are created by $ \pm 1$ embedding on the horizontal or vertical components of MVs. If not specified, all the following experiments in this section are conducted under the same setting. We can see from Fig. \ref{fig:PMV_bitrate} that the change rates of PMVs and bit-rate both increase linearly with increasing the change rate of MVs, demonstrating that the PMV and bit-rate (also the rate-distortion cost) are indeed affected by the steganographic embedding on MVs. Another observation is that the change rates of PMVs and bit-rate are lower than that of MVs, because the PMV is the median of three neighboring MVs, and modifying the neighboring MVs does not necess-arily cause a change in the corresponding PMV. For the bit-rate, it has a ladder-like change with the MVD (see Fig. 8 in Supplementary), so its change rate is lower than that of PMVs.


\subsection{Four cases of Estimation for Local optimality}  \label{sec:glo_b}	
The local optimality of MVs is disturbed during steganographic embedding at the video encoder side, but has to be estimated for steganalysis at the video decoder side, so the estimation of local optimality is affected by the quantization distortion \cite{Wang2014AoSO}. We verify that apart from the quantization distortion, the change of PMV will also affect the estimation of local optimality.

Since the rate-distortion cost is jointly determined by PMV and MV, the local optimality of a MV can be estimated in following four cases:
\begin{itemize}
\item Case 1: both the PMV and the MV are not changed.
\item Case 2: the PMV is not changed, but the MV is changed.
\item Case 3: the PMV is changed, but the MV is not changed.
\item Case 4: both the PMV and the MV are changed.
\end{itemize}

In the first two cases, the PMVs are not changed during the embedding, which is considered by previous method \cite{Zhang2017NPE}. While in the last two cases, the PMVs are changed due to the changes of other MVs. The empirical probabilities of occurrences of four cases in stego videos are listed in Table \ref{tab:four_cases_occur}.

\begin{table}[t]
  \newcommand{\tabincell}[2]{\begin{tabular}{@{}#1@{}}#2\end{tabular}}
  \centering
  \caption{The empirical probabilities of occurrences of four cases in stego videos}
  \vspace{-4pt}
  \label{tab:four_cases_occur}
  \centering
  \begin{tabular}{cccccccc}
  \toprule
  \multirow{2}{*}[-2pt]{\tabincell{c}{QP}} & \multirow{2}{*}[-2pt]{\tabincell{c}{Four\\Cases}} & \multicolumn{6}{c}{Change Rate of MVs} \\
  \cmidrule{3-8}
    & & 0.05 & 0.1 & 0.2 & 0.3 & 0.4 & 0.5 \\
    \midrule
    \multirow{4}{*}{15} &
      Case 1 & 0.9110 & 0.8262 & 0.6695 & 0.5294 & 0.4062 & 0.2994 \\
    & Case 2 & 0.0480 & 0.0918 & 0.1675 & 0.2271 & 0.2709 & 0.2996 \\
    & Case 3 & 0.0390 & 0.0738 & 0.1305 & 0.1705 & 0.1938 & 0.2005 \\
    & Case 4 & 0.0020 & 0.0082 & 0.0325 & 0.0729 & 0.1291 & 0.2005 \\
    \midrule
    \multirow{4}{*}{25} &
      Case 1 & 0.9127 & 0.8293 & 0.6739 & 0.5343 & 0.4107 & 0.3024 \\
    & Case 2 & 0.0481 & 0.0923 & 0.1687 & 0.2294 & 0.2741 & 0.3033 \\
    & Case 3 & 0.0373 & 0.0707 & 0.1261 & 0.1656 & 0.1893 & 0.1973 \\
    & Case 4 & 0.0019 & 0.0078 & 0.0313 & 0.0707 & 0.1259 & 0.1970 \\
    \bottomrule
    \end{tabular}
\end{table}

\begin{table}[t]
  \newcommand{\tabincell}[2]{\begin{tabular}{@{}#1@{}}#2\end{tabular}}
  \centering
  \caption{The empirical probabilities of occurrences of local optimality for four cases in stego videos}
  \vspace{-4pt}
  \label{tab:four_cases_opt}
  \centering
  \begin{tabular}{cccccccc}
  \toprule
  \multirow{2}{*}[-2pt]{\tabincell{c}{QP}} & \multirow{2}{*}[-2pt]{\tabincell{c}{Four\\Cases}} & \multicolumn{6}{c}{Change Rate of MVs} \\
  \cmidrule{3-8}
    & & 0.05 & 0.1 & 0.2 & 0.3 & 0.4 & 0.5 \\
    \midrule
    \multirow{4}{*}{15} &
      Case 1 & 0.7608 & 0.7611 & 0.7635 & 0.7650 & 0.7670 & 0.7680 \\
    & Case 2 & 0.3301 & 0.3290 & 0.3302 & 0.3302 & 0.3310 & 0.3308 \\
    & Case 3 & 0.7218 & 0.7233 & 0.7260 & 0.7279 & 0.7304 & 0.7322 \\
    & Case 4 & 0.3366 & 0.3357 & 0.3370 & 0.3373 & 0.3379 & 0.3380 \\
    \midrule
    \multirow{4}{*}{25} &
      Case 1 & 0.8568 & 0.8581 & 0.8605 & 0.8636 & 0.8656 & 0.8682 \\
    & Case 2 & 0.4881 & 0.4888 & 0.4912 & 0.4938 & 0.4950 & 0.4980 \\
    & Case 3 & 0.8060 & 0.8072 & 0.8105 & 0.8146 & 0.8173 & 0.8208 \\
    & Case 4 & 0.4788 & 0.4844 & 0.4818 & 0.4852 & 0.4875 & 0.4900 \\
    \bottomrule
    \end{tabular}
\end{table}

As shown in Table \ref{tab:four_cases_occur}, for a given change rate of MVs, the empirical probabilities of each case are very close under different QPs, indicating that the proportions of four cases are not influenced by the compression degree. When the change rate increases, the proportion of Case 1 decreases gradually, while the proportions of the other three cases all increase. In particular, the empirical probability of Case 4 increases more significantly, because the interaction among neighboring MVs is more prominent at high change rates.

To investigate the embedding effect on the local optimality for different cases, we report in Table \ref{tab:four_cases_opt} the empirical probabilities of occurrences of local optimality for four cases in stego videos. This experiment is simulated at the video encoder side, from which we can check the values of PMVs and MVs before and after embedding, and the rate-distortion costs are calculated using the updated PMVs after embedding, which matches the situation at the video decoder side.

By comparing Case 1 and Case 2 in Table \ref{tab:four_cases_opt}, we observe a large probability gap between them. This is because the Case 1 corresponds to the situation in a cover video where both the PMVs and the MVs remain unchanged, and naturally has high probabilities of local optimality. The Case 2 is an ideal situation where only the MVs are changed and the rate-distortion costs can be obtained free from the influence of PMVs, so it has low probabilities of local optimality.

\begin{figure}[t]
  \centering
  \includegraphics[width=0.936\columnwidth]{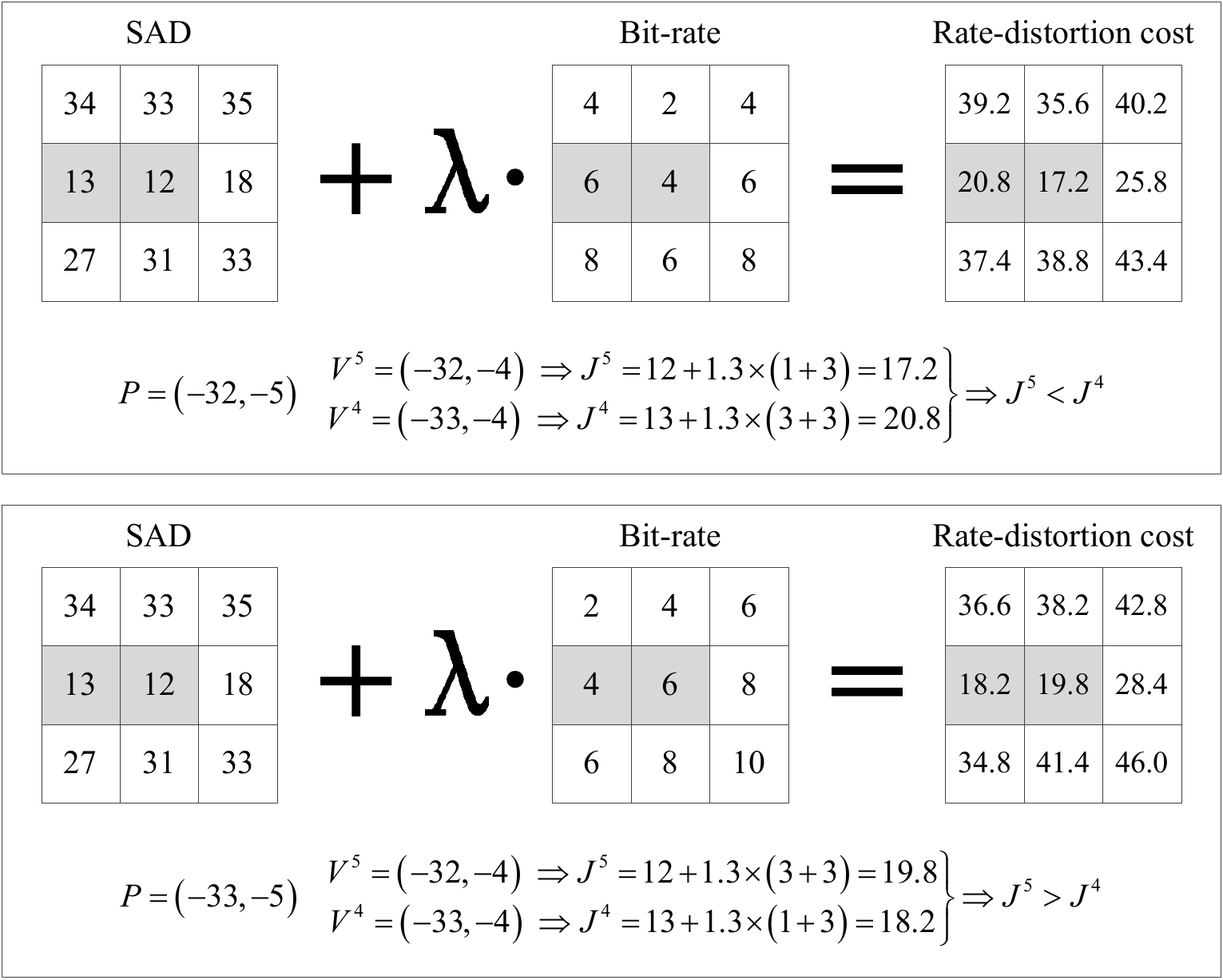}
  \caption{The changes of SAD, bit-rate and rate-distortion cost for Case 3.}
  \vspace{-4pt}
  \label{fig:Example_case3}
\end{figure}

\begin{figure}[t]
  \centering
  \subfloat[]{\includegraphics[width=0.42\columnwidth]{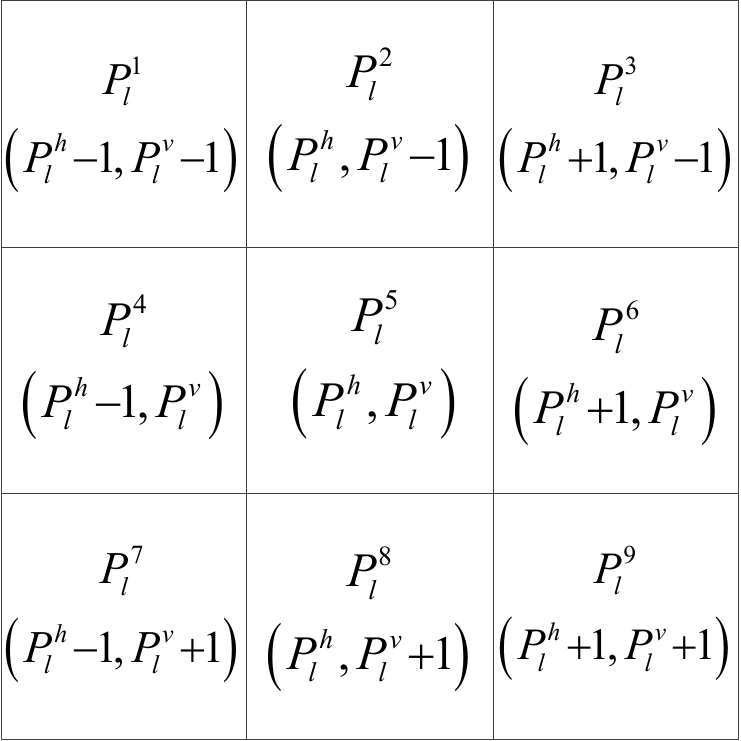}\label{fig:Spa_struc_a}}
  \hfil
  \subfloat[]{\includegraphics[width=0.42\columnwidth]{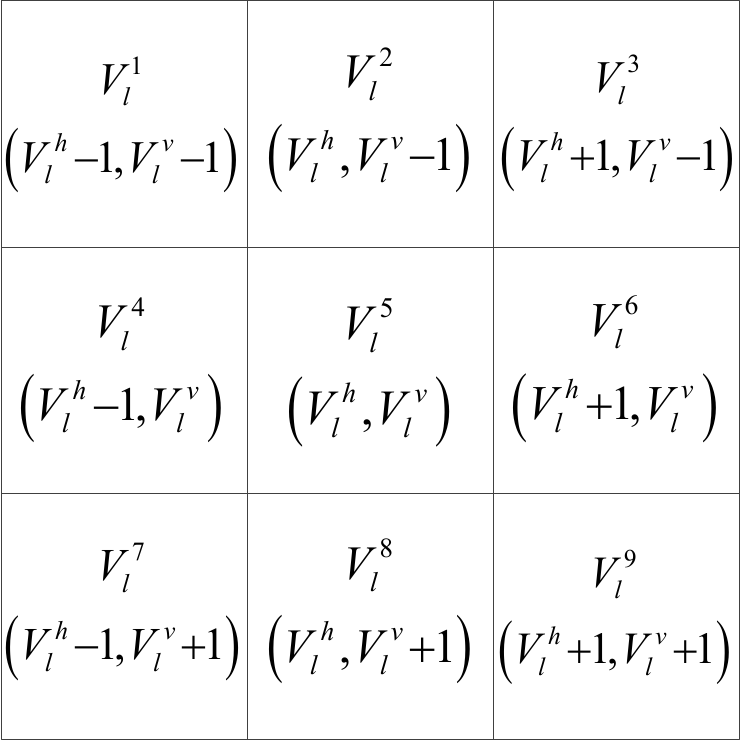}\label{fig:Spa_struc_b}}
  \vfil
  \subfloat[]{\includegraphics[width=0.42\columnwidth]{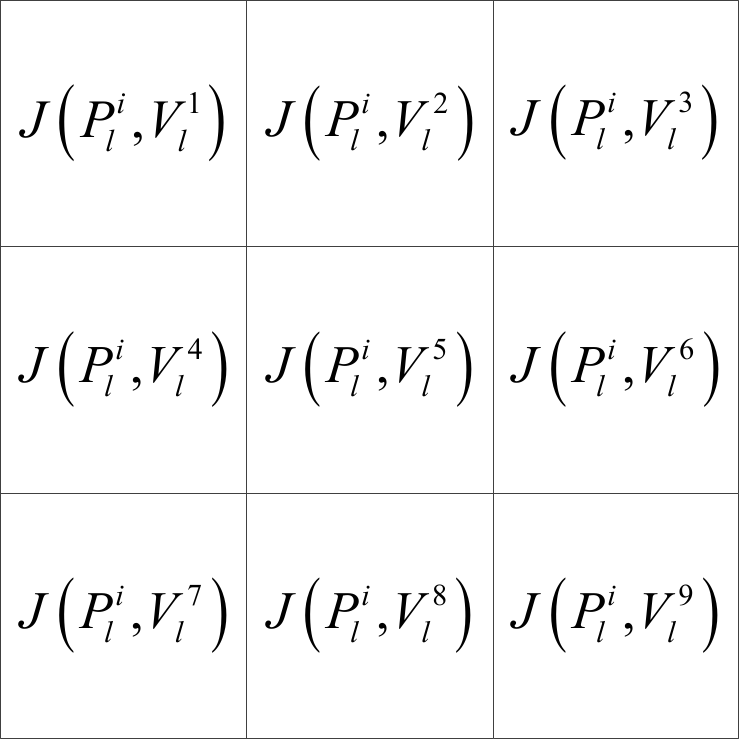}\label{fig:Spa_struc_c}}
  \hfil
  \subfloat[]{\includegraphics[width=0.42\columnwidth]{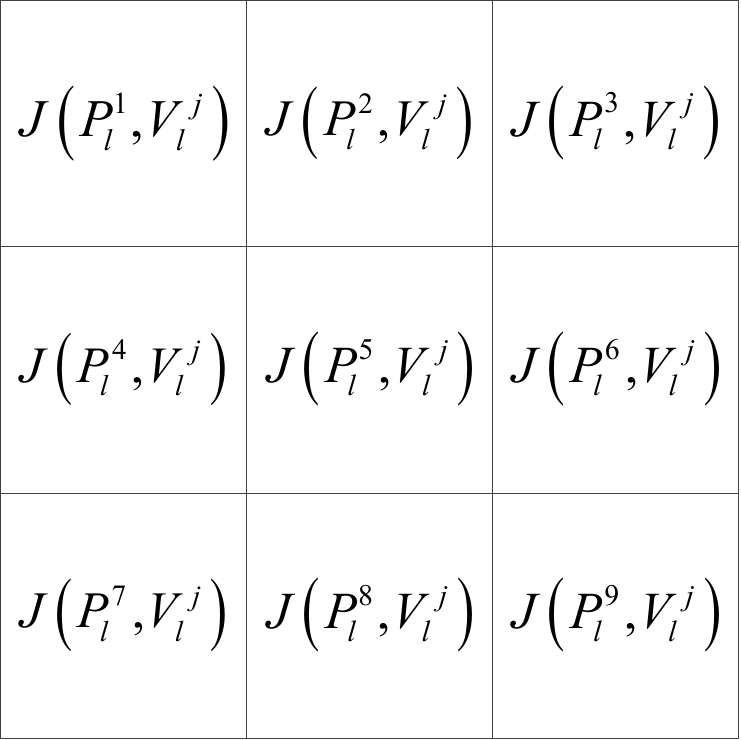}\label{fig:Spa_struc_d}}
  \caption{The relative spatial positions of (a) PMV set $\Omega ({P_l})$, (b) MV set $\Omega ({V_l})$ and (c) (d) their rate-distortion costs.}
  \label{fig:Spa_struc}
\end{figure}

For the Case 3, although the MVs are not changed, the local optimality can still be destroyed by the changes of PMVs. Because the changed PMV may alter the rate-distortion cost, deviating the MV from locally optimal to locally non-optimal. This process is illustrated by the example in Fig. \ref{fig:Example_case3}, showing the changes of SAD, bit-rate and rate-distortion cost when only the PMV is modified. The results in Fig. \ref{fig:Example_case3} are obtained from a coding block (frame No. 12, block No. 44) in video sequence ``bus'' in DB1. The $P$ denotes the PMV, the ${V^5}$ is the original MV located \mbox{in the center of the $3 \times 3$ neighborhood} (see the $3 \times 3$  structure in Fig. \ref{fig:Spa_struc}\subref{fig:Spa_struc_b}), ${V^4}$ is the modified MV by making embedding change ``$-1$'' on the horizontal component of ${V^5}$, and ${J^4}$ and ${J^5}$ are the rate-distortion costs corresponding to ${V^4}$ and ${V^5}$. The video sequence is H.264/AVC compressed with QP 15, so the Lagrangian multiplier is $\lambda  = 1.3$ (see the definition of $\lambda$ in (\ref{eq:rdo})).

As can be seen from Fig. \ref{fig:Example_case3}, before changing the PMV $P$, the ${V^5}$ has a local optimality, \ie, ${J^5} = \min \left\{ {{J^k}} \right\},k = 1, \cdots ,9$. After changing the PMV from $P = \left( { - 32, - 5} \right)$ to $P = \left( { - 33, - 5} \right)$, the SADs remain unchanged, but the MVDs together with bit-rates and rate-distortion costs in the $3 \times 3$ neighborhood are all changed. Taking ${V^4}$ and ${V^5}$ as an example, since ${V^5}$ and $P$ have the same modification directions (both $-1$ on the horizontal components), the ${V^4}$ changed from ${V^5}$ tends to have a smaller magnitude of MVD, and thus the ${V^4}$ has a lower bit-rate and a lower rate-distortion cost than ${V^5}$. Finally, when rechecking local optimality in the $3 \times 3$ neighborhood, the ${V^5}$ is no longer locally optimal.

As for the Case 4, it is the combination of Case 2 and Case 3. The superposition of changing both the PMV and the MV may shift the MV from locally optimal to locally non-optimal, or even shift the MV from locally non-optimal (caused by quantization distortion) to locally optimal. These two types of changes for local optimality occur almost equally by our experimental observation, so the overall empirical probabilities of Case 2 and Case 4 are very close. For brevity and simplicity, we do not provide detail examples like Fig. \ref{fig:Example_case3}.

From the explanations of Case 3 and Case 4, we observe that the changes of the PMVs may deviate the estimation of local optimality for tested MVs. This deviation is sometimes beneficial to video steganalysis (\eg, the changed PMV make the unchanged MV become locally non-optimal), but sometimes it is unfavorable to steganalysis (the changed PMV and the changed MV may counteract with each other, obtaining a locally optimal MV and thus reducing the embedding traces). Therefore, we draw following two conclusions:
\begin{enumerate}[ 1)]
\item To increase the accuracy of estimation for local optimality, we suggest checking the local optimality by considering the dynamics of PMVs.
\item To improve the diversity of local optimality based features, we suggest checking the local optimality from the view of both MVs and PMVs.
\end{enumerate}

In the following two subsections, we will generalize the local optimality based on the two conclusions and demonstrate their effectiveness.

%
\begin{figure*}[t]
  \centering
  \subfloat[]{\includegraphics[width=0.47\textwidth]{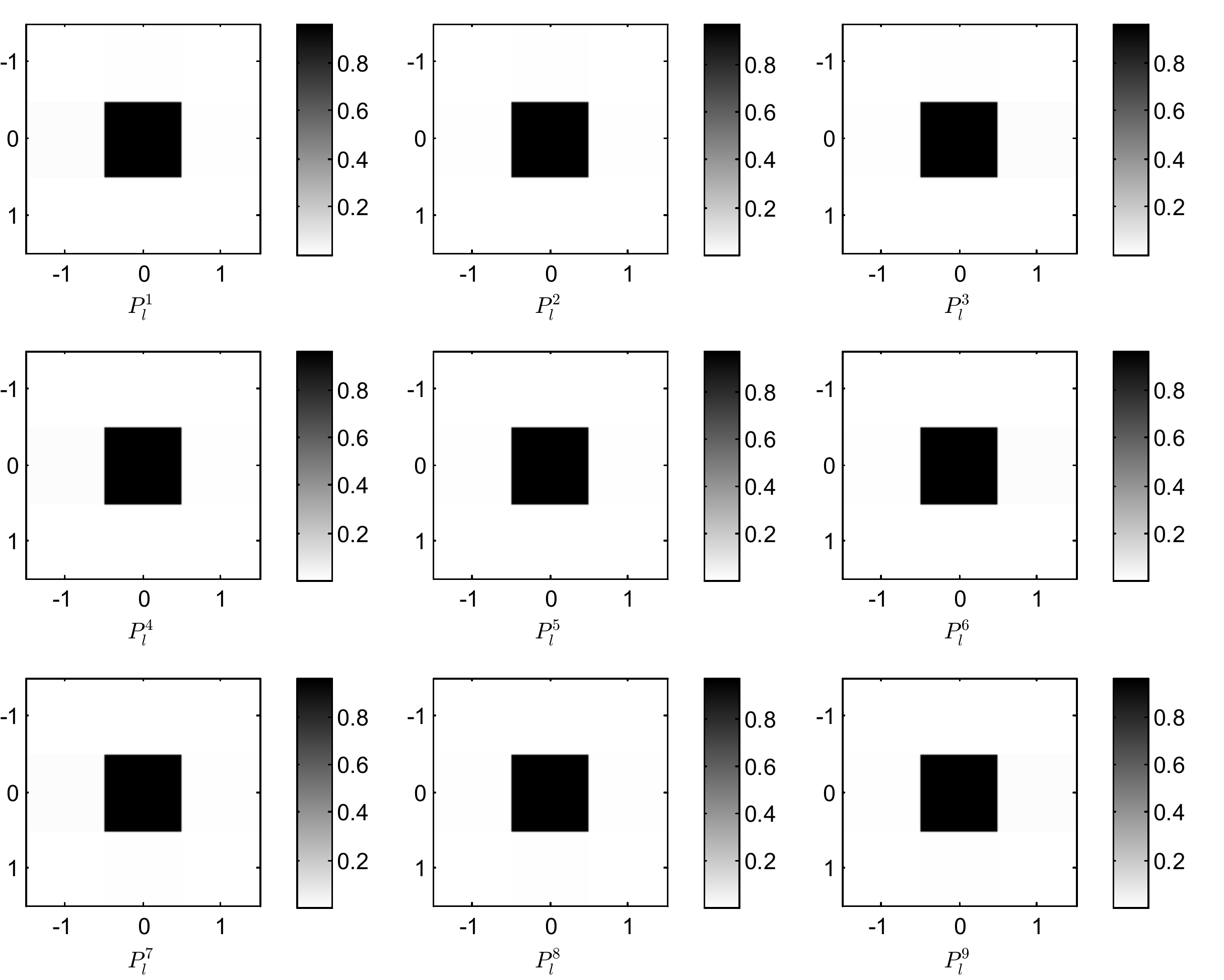}\label{fig:GLO_MV_prob_a}}
  \hfil
  \subfloat[]{\includegraphics[width=0.47\textwidth]{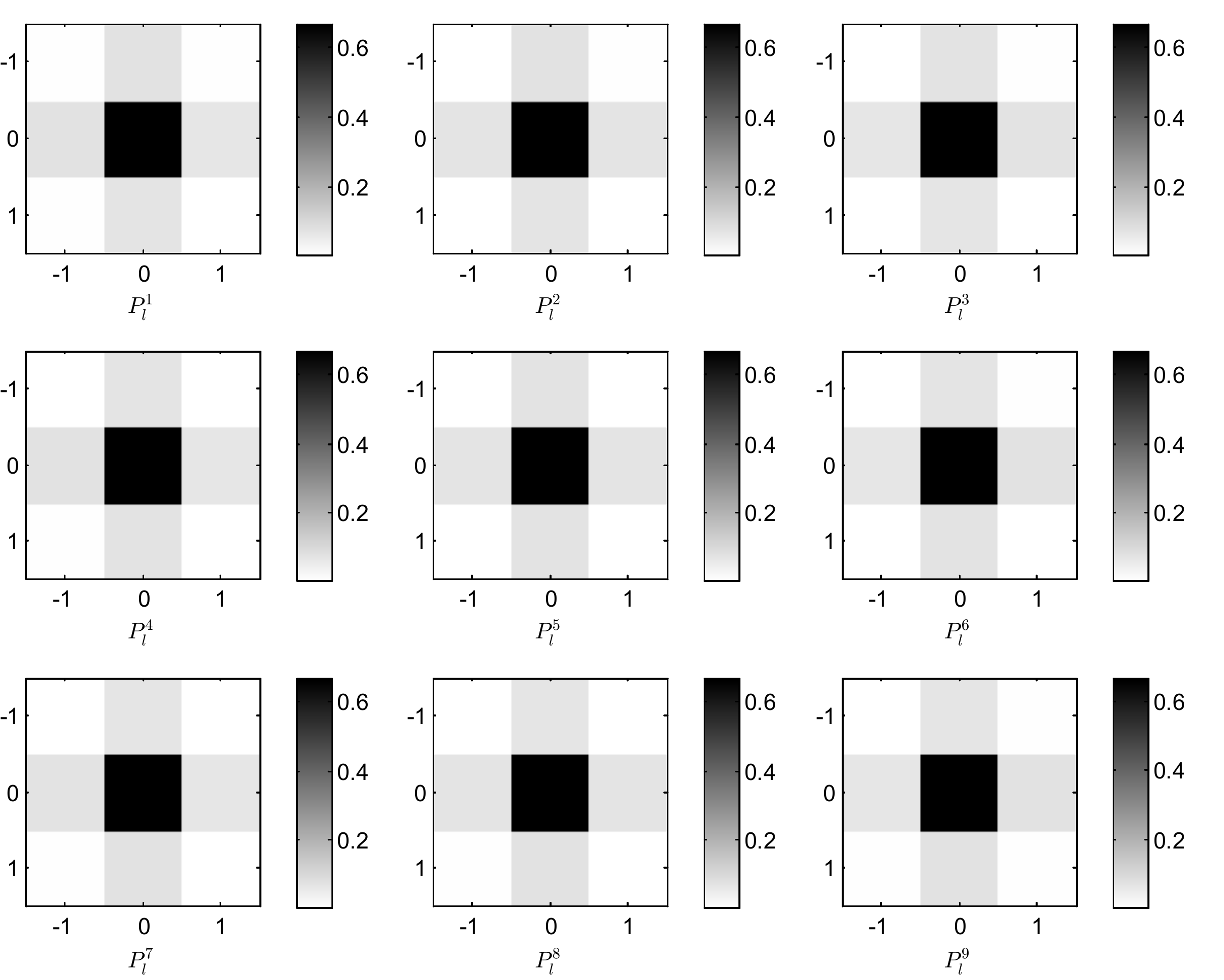}\label{fig:GLO_MV_prob_b}}
  \caption{Empirical probability $\Pr (J(P_l^i,V_l^j) = {J_{\min }}(P_l^i,\Omega ({V_l})))$ for (a) cover and (b) stego videos.}
  \label{fig:GLO_MV_prob}
\end{figure*}

\begin{figure*}[t]
  \centering
  \subfloat[]{\includegraphics[width=0.47\textwidth]{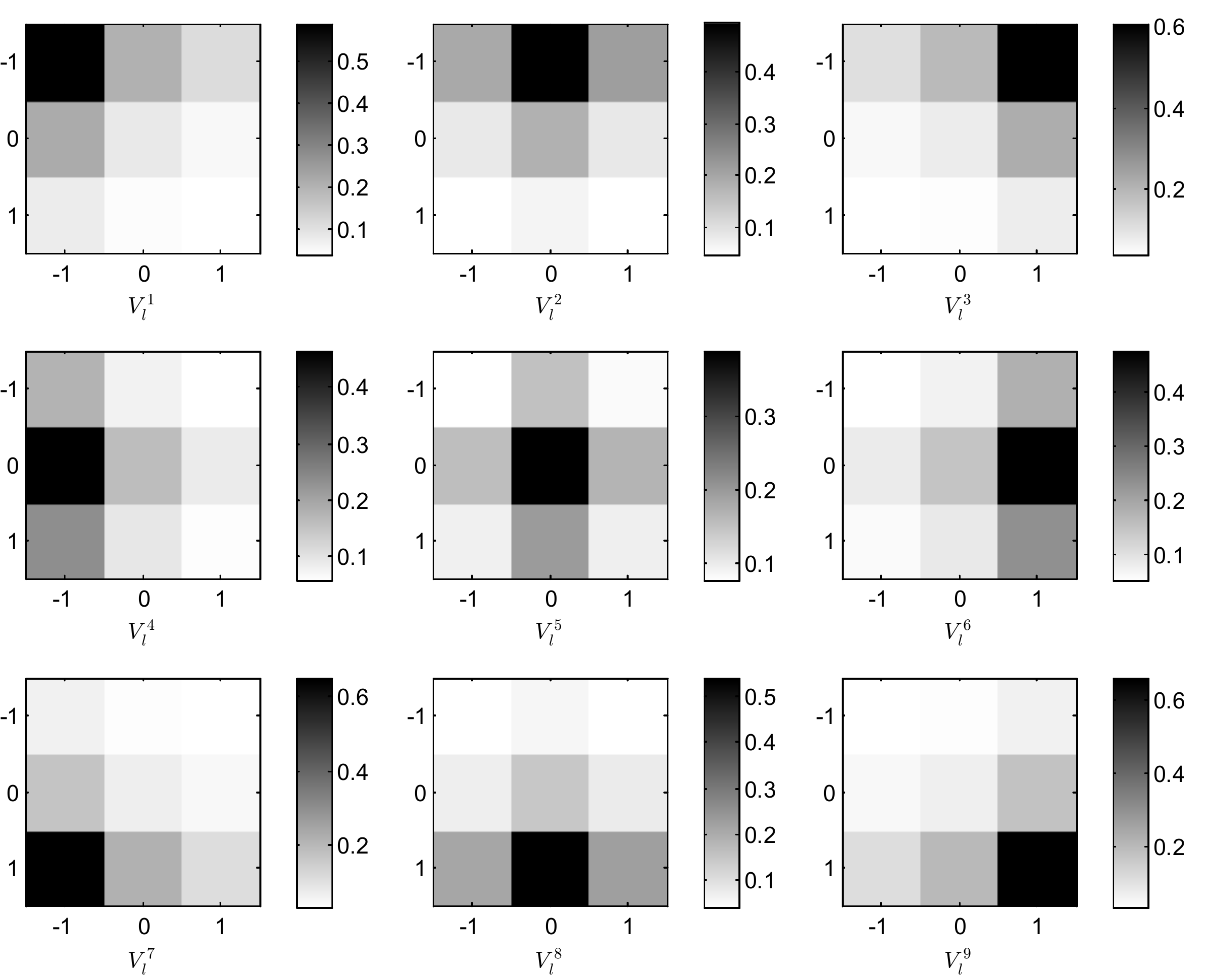}\label{fig:GLO_PMV_prob_a}}
  \hfil
  \subfloat[]{\includegraphics[width=0.47\textwidth]{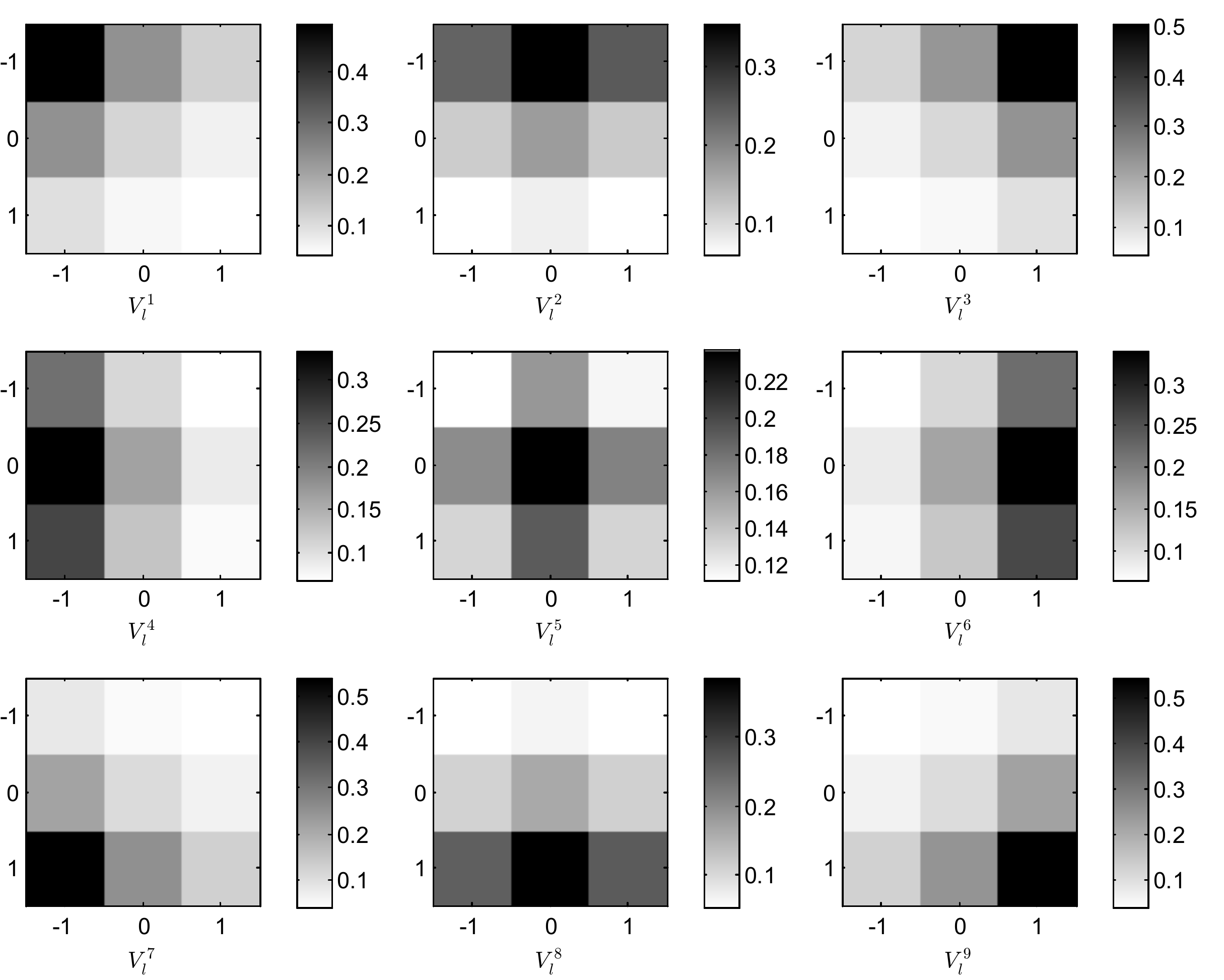}\label{fig:GLO_PMV_prob_b}}
  \caption{Empirical probability $\Pr (J(P_l^i,V_l^j) = {J_{\min }}(\Omega ({P_l}),V_l^j))$ for (a) cover and (b) stego videos.}
  \label{fig:GLO_PMV_prob}
\end{figure*}

\subsection{Generalized Local Optimality of MVs}  \label{sec:glo_c}

According to the first conclusion, the dynamics of PMVs should be considered to estimate the local optimality of MVs. However, the steganalyst at the video decoder side does not know whether or not the PMVs have been changed or the way they were changed. So it is impossible to estimate the local optimality of MVs by predetermining the status of PMVs.

Inspired by the local optimality estimation that considers the MVs with all potential modifications, we estimate the local optimality of MVs under the conditions of all potential modifications of PMVs.

We define the generalized local optimality of a MV as the optimality that the corresponding MV has the lowest rate-distortion cost under various possible modifications of the PMV. Since larger modification magnitudes of MVs lead to greater impact on video compression efficiency and steganographic security, practical MV based steganography often limits the embedding changes to $ \pm 1$ \cite{Xu2006MV, Aly2011MV, Cao2011PME, Yao2015MV, Cao2015PMEopt, Zhu2018UED}. Following the practice in \cite{Wang2014AoSO, Ren2014SPOM, Zhang2017NPE}, we also consider the $ \pm 1$ embedding on MVs. Let the $l$-th MV in a frame be denoted by ${V_l} = (V_l^h,V_l^v)$, where $V_l^h$ and $V_l^v$ are the horizontal and vertical components of ${V_l}$. A set of candidate MVs which are potentially modified MVs is defined as
\begin{equation}
  \Omega ({V_l}) = \left\{ {V_l^h \!-\! 1,V_l^h,V_l^h \!+\! 1} \right\} \times \left\{ {V_l^v \!-\! 1,V_l^v,V_l^v \!+\! 1} \right\}.
  \label{eq:mv_set}
\end{equation}

Let the PMV of the $l$-th MV be denoted by ${P_l} = (P_l^h,P_l^v)$, where $P_l^h$ and $P_l^v$ are the horizontal and vertical components of $P_l$. Similar to (\ref{eq:mv_set}), a set of candidate PMVs which are potentially modified PMVs is defined as
\begin{equation}
  \Omega ({P_l}) = \left\{ {P_l^h \!-\! 1,P_l^h,P_l^h \!+\! 1} \right\} \times \left\{ {P_l^v \!-\! 1,P_l^v,P_l^v \!+\! 1} \right\}.
  \label{eq:pmv_set}
\end{equation}

The relative spatial positions of $\Omega ({P_l})$ and $\Omega ({V_l})$ are $3 \times 3$ neighborhoods as shown in Fig. \ref{fig:Spa_struc}\subref{fig:Spa_struc_a} and \subref{fig:Spa_struc_b}. Let $J(P_l^i,V_l^j)$ denotes the rate-distortion cost determined by $P_l^i$ and $V_l^j$. Given a PMV $P_l^i \in \Omega ({P_l})$, the relative spatial positions of $\left\{ {\left. {J(P_l^i,V_l^j)} \right|j = 1, \cdots ,9} \right\}$ is shown in Fig. \ref{fig:Spa_struc}\subref{fig:Spa_struc_c}, and the empirical probability that the $V_l^j \in \Omega ({V_l})$ has a local optimality is denoted by $\Pr (J(P_l^i,V_l^j) = {J_{\min }}(P_l^i,\Omega ({V_l})))$, where ${J_{\min }}(P_l^i,\Omega ({V_l})) = \min \left\{ {\left. {J(P_l^i,V_l^j)} \right|j = 1, \cdots ,9} \right\}$. The comparison of $\Pr (J(P_l^i,V_l^j) = {J_{\min }}(P_l^i,\Omega ({V_l})))$ for cover and stego videos are shown in Fig. \ref{fig:GLO_MV_prob}.
The cover and stego videos are the same as those mentioned in Section \ref{sec:glo_a}.
The videos are compressed with QP 15, and the local optimality is measured by the SAD based rate-distortion cost. To highlight the graphical comparison results, a large change rate of 0.5 for MVs is used.

As seen from Fig. \ref{fig:GLO_MV_prob}, in stego videos, the $\Pr (J(P_l^i,V_l^5) = {J_{\min }}(P_l^i,\Omega ({V_l})))$ is decreased and the $\Pr (J(P_l^i,V_l^j) = {J_{\min }}(P_l^i,\Omega ({V_l}))),\;j \ne 5$ are all increased, verifying that the steganographic embedding destroys the local optimality of MVs. The above phenomenon is observed for all $P_l^i \in \Omega ({P_l})$, this demonstrates that the dynamics of PMVs should indeed be used to estimate the local optimality of MVs (the previous work \cite{Zhang2017NPE} regards all $P_l^i \in \Omega ({P_l})$ as $P_l^5$.

\subsection{Generalized local optimality of PMVs}  \label{sec:glo_d}

The MV based steganography directly modifies the MVs, so the local optimality in existing methods \cite{Wang2014AoSO, Ren2014SPOM, Zhang2017NPE} is naturally analyzed from the view of MVs. Since the rate-distortion cost is jointly determined by the MV and the PMV, the local optimality can also be explained from the view of the PMV, as illustrated by the example of Case 3 in Fig. \ref{fig:Example_case3}. Specifically, for a block whose MV is not changed, the corresponding rate-distortion cost will depend on the PMV. If the PMV is also unchanged, we could say that the PMV has a local optimality. We could also say that the PMV is locally non-optimal when its rate-distortion cost is not minimal due to the change of the PMV. For the varied MV, the local optimality can also be checked in terms of the PMV, as did in Section \ref{sec:glo_c}.

Similarly, we define the generalized local optimality of a PMV as the optimality that the corresponding PMV has the lowest rate-distortion cost under various possible modifications of the MV. Given a MV $V_l^j \in \Omega ({V_l})$, the relative spatial positions of $\left\{ {\left. {J(P_l^i,V_l^j)} \right|i = 1, \cdots ,9} \right\}$ is shown in Fig. \ref{fig:Spa_struc}\subref{fig:Spa_struc_d}, and the empirical probability that the $P_l^i \in \Omega ({P_l})$ has a local optimality is denoted by $\Pr (J(P_l^i,V_l^j) = {J_{\min }}(\Omega ({P_l}),V_l^j))$, where ${J_{\min }}(\Omega ({P_l}),V_l^j) = \min \left\{ {\left. {J(P_l^i,V_l^j)} \right|i = 1, \cdots ,9} \right\}$. The comparison of $\Pr (J(P_l^i,V_l^j) = {J_{\min }}(\Omega ({P_l}),V_l^j))$ for cover and stego videos are shown in Fig. \ref{fig:GLO_PMV_prob}, and the experimental settings are the same as those for Fig. \ref{fig:GLO_MV_prob}.

We can see from Fig. \ref{fig:GLO_PMV_prob} that the highest empirical probabilities in the $3 \times 3$ neighborhoods are all tilted to the direction of embedding changes. Taking $\Pr (J(P_l^i,V_l^4) = {J_{\min }}(\Omega ({P_l}),V_l^4))$ as an example, the $V_l^4$ is obtained by making embedding change ``$ - 1$'' on the horizontal component of ${V_l}$ (\ie., $V_l^5$), and the $P_l^4$ that has the same modification direction as $V_l^4$ is prone to have a smaller magnitude of MVD (\ie, lower bit-rate), so $\Pr (J(P_l^4,V_l^4) = {J_{\min }}(\Omega ({P_l}),V_l^4))$ has the maximum value. When comparing the cover videos and the stego videos, we also see that the steganographic embedding on MVs disturbs the local optimality of PMVs. For example, the $\Pr (J(P_l^i,V_l^j) = {J_{\min }}(\Omega ({P_l}),V_l^j)),\;i = j$ in cover videos is larger than that in stego videos, while the $\Pr (J(P_l^i,V_l^j) = {J_{\min }}(\Omega ({P_l}),V_l^j)),\;i \ne j$ in cover videos is lower than that in stego videos (see the ranges of color bars in Fig. \ref{fig:GLO_PMV_prob}). This is clear evidence that the generalized local optimality of PMVs also has a strong statistical discriminability, which can be used to design steganalytic features.

\section{The Proposed Features}  \label{sec:fea}

\subsection{Feature Construction}  \label{sec:fea_a}


For each MV ${V_l} = (V_l^h,V_l^v)$ and its PMV ${P_l} = (P_l^h,P_l^v)$, the following steps of preparation are needed for feature construction.
\begin{enumerate}[ 1)]
\item Obtain the MV set $\Omega ({V_l})$ and the PMV set $\Omega ({P_l})$ based on (\ref{eq:mv_set}) and (\ref{eq:pmv_set}).

\item For each combination of $P_l^i \in \Omega ({P_l})$ and $V_l^j \in \Omega ({V_l})$, calculate its rate-distortion cost ${J^D}(P_l^i,V_l^j)$, where $D \in \left\{ {{\rm{SAD}},{\rm{SATD}}} \right\}$ denotes the rate-distortion cost based on SAD or SATD.

%

\item For each fixed $P_l^i \in \Omega ({P_l})$, check the rate-distortion costs in $\left\{ {\left. {{J^D}(P_l^i,V_l^j)} \right|j = 1, \cdots ,9} \right\}$, and determine the minimum one as $J_{\min }^D(P_l^i,\Omega ({V_l}))$.

\item For each fixed $V_l^j \in \Omega ({V_l})$, check the rate-distortion costs in $\left\{ {\left. {{J^D}(P_l^i,V_l^j)} \right|i = 1, \cdots ,9} \right\}$, and determine the minimum one as $J_{\min }^D(\Omega ({P_l}),V_l^j)$.
\end{enumerate}

The steganalytic features based on the generalized local optimality of MVs have two types. The first type corresponds to the probability that ${J^D}(P_l^i,V_l^j)$ equals $J_{\min }^D(P_l^i,\Omega ({V_l}))$ with a given $P_l^i$, $V_l^j$ and $D$, and is defined by
\begin{equation}
  \begin{aligned}
    f_{1,k}^D =& \Pr \left( {{J^D}(P_l^i,V_l^j) = J_{\min }^D(P_l^i,\Omega ({V_l}))} \right)\\
    =& \frac{1}{L}\sum\limits_{l = 1}^L {\delta \left( {{J^D}(P_l^i,V_l^j),J_{\min }^D(P_l^i,\Omega ({V_l}))} \right)} \\
    &( i,j = 1, \cdots ,9,\;k = 9 \cdot (i - 1) + j, \\
    &D \in \left\{ {{\rm{SAD}},{\rm{SATD}}} \right\} )
  \end{aligned}
  \label{eq:glo_mv_fea1}
\end{equation}
where $L$ is the number of MVs (PMVs) in a video frame, and $\delta \left( {x,y} \right) = 1$ if $x = y$ and 0 otherwise.

The second type is the exponentially magnified relative difference between ${J^D}(P_l^i,V_l^j)$ and $J_{\min }^D(P_l^i,\Omega ({V_l}))$:
\begin{equation}
  \begin{aligned}
    f_{2,k}^D =& \frac{1}{Z}\sum\limits_{l = 1}^L {\exp \left\{ {\frac{{\left| {{J^D}(P_l^i,{V_l^5}) - {J^D}(P_l^i,V_l^j)} \right|}}{{{J^D}(P_l^i,{V_l^5})}}} \right\}} \\
    &\cdot \delta \left( {{J^D}(P_l^i,V_l^j),J_{\min }^D(P_l^i,\Omega ({V_l}))} \right)\\
    &( i,j = 1, \cdots ,9,\;k = 9 \cdot (i - 1) + j,\\
    &D \in \left\{ {{\rm{SAD}},{\rm{SATD}}} \right\} )\\
    Z =& \sum\limits_{i = 1}^9 {\sum\limits_{l = 1}^L {\exp \left\{ {\frac{{\left| {{J^D}(P_l^i,{V_l^5}) - {J^D}(P_l^i,V_l^j)} \right|}}{{{J^D}(P_l^i,{V_l^5})}}} \right\}} } \\
    &\cdot \delta \left( {{J^D}(P_l^i,V_l^j),J_{\min }^D(P_l^i,\Omega ({V_l}))} \right)
\end{aligned}
  \label{eq:glo_mv_fea2}
\end{equation}

The steganalytic features based on the generalized local optimality of PMVs also have two types, and the first type corresponds to the probability that ${J^D}(P_l^i,V_l^j)$ equals $J_{\min }^D(\Omega ({P_l}),V_l^j)$ with a given $V_l^j$, $P_l^i$ and $D$:
\begin{equation}
  \begin{aligned}
    f_{3,k}^D =& \Pr \left( {{J^D}(P_l^i,V_l^j) = J_{\min }^D(\Omega ({P_l}),V_l^j)} \right)\\
    =& \frac{1}{L}\sum\limits_{l = 1}^L {\delta \left( {{J^D}(P_l^i,V_l^j),J_{\min }^D(\Omega ({P_l}),V_l^j)} \right)} \\
    &\left( {i,j = 1, \cdots ,9, k = 9 \cdot (j - 1) + i, D \!\in\! \left\{ {{\rm{SAD}}} \right\}} \right)
\end{aligned}
  \label{eq:glo_pmv_fea1}
\end{equation}

The second type is also the exponentially magnified relative difference between ${J^D}(P_l^i,V_l^j)$ and $J_{\min }^D(\Omega ({P_l}),V_l^j)$:
\begin{equation}
  \begin{aligned}
    f_{4,k}^D =& \frac{1}{Z}\sum\limits_{l = 1}^L {\exp \left\{ {\frac{{\left| {{J^D}({P_l^5},V_l^j) - {J^D}(P_l^i,V_l^j)} \right|}}{{{J^D}({P_l^5},V_l^j)}}} \right\}} \\
    &\cdot \delta \left( {{J^D}(P_l^i,V_l^j),J_{\min }^D(\Omega ({P_l}),V_l^j)} \right)\\
    &\left( {i,j = 1, \cdots ,9, k = 9 \cdot (j - 1) + i, D \!\in\! \left\{ {{\rm{SAD}}} \right\}} \right)\\
    Z =& \sum\limits_{i = 1}^9 {\sum\limits_{l = 1}^L {\exp \left\{ {\frac{{\left| {{J^D}({P_l^5},V_l^j) - {J^D}(P_l^i,V_l^j)} \right|}}{{{J^D}({P_l^5},V_l^j)}}} \right\}} } \\
    &\cdot \delta \left( {{J^D}(P_l^i,V_l^j),J_{\min }^D(\Omega ({P_l}),V_l^j)} \right)
\end{aligned}
  \label{eq:glo_pmv_fea2}
\end{equation}

For the generalized local optimality of MVs, the features have two types, and each of them has two types of rate-distortion costs (SAD based or SATD based), nine $P_l^i$ and nine $V_l^j$, so the feature dimension is $2 \times 2 \times \left( {9 \times 9} \right) = 324$.

For the generalized local optimality of PMVs, the visual distortion is only controlled by the MV, and is irrelevant to the PMV (\ie., $f_{3,k}^{{\rm{SAD}}} = f_{3,k}^{{\rm{SATD}}}$), so only the SAD based rate-distortion cost is used for the first type of features. As for the ${J^D}(P_l^i,V_l^j)$ and $J_{\min }^D(\Omega ({P_l}),V_l^j)$ in the second type of features, although their exponentially relative difference is related to the SAD and SATD, their values are very close especially for high QPs. Besides, we also demonstrate in Section V-B that the steganalysis performance of $f_{4,k}^{{\rm{SAD}}}$, $f_{4,k}^{{\rm{SATD}}}$ and their combination is almost the same. So only the SAD based rate-distortion cost is used for the second type of features. As a result, the feature dimension is $2 \times \left( {9 \times 9} \right) = 162$.

The proposed features are referred to as generalized local optimality (GLO) features, and the sub-features based on the generalized local optimality of MVs and PMVs are referred to as GLO-MV and GLO-PMV, respectively.

\subsection{Feature Symmetrization} \label{sec:fea_b}

The proposed features have a much higher dimensionality than the previous local optimality based features \cite{Wang2014AoSO, Ren2014SPOM, Zhang2017NPE}. To explore whether there is redundancy in the proposed features, this subsection studies feature symmetrization to reduce feature dimension.

As observed from Fig. \ref{fig:GLO_MV_prob} and Fig. \ref{fig:GLO_PMV_prob}, the empirical probability distributions exhibit specific symmetrical patterns, such as the cross symmetry for the generalized local optimality of MVs, and the directional symmetry along embedding changes for the generalized local optimality of PMVs.

%
According to the modification type and the empirical probability distributions in Fig. \ref{fig:GLO_MV_prob}, the feature symmetrization process for the GLO-MV is performed as follows:
\begin{enumerate}[ 1)]
\item Divide the PMV set $\Omega ({P_l})$ into \textbf{three} subsets ${\cal P}_l^1 = \left\{ {P_l^1,P_l^3,P_l^7,P_l^9} \right\}$, ${\cal P}_l^2 = \left\{ {P_l^2,P_l^4,P_l^6,P_l^8} \right\}$ and ${\cal P}_l^3 = \left\{ {P_l^5} \right\}$, and all PMVs in each subset are regarded as the same PMV.

\item Divide the MV set $\Omega ({V_l})$ into \textbf{three} subsets ${\cal V}_l^1 = \left\{ {V_l^1,V_l^3,V_l^7,V_l^9} \right\}$, ${\cal V}_l^2 = \left\{ {V_l^2,V_l^4,V_l^6,V_l^8} \right\}$ and ${\cal V}_l^3 = \left\{ {V_l^5} \right\}$, and all MVs in each subset are regarded as the same MV.

\item Combine each PMV subset and each MV subset, and extract the features based on (\ref{eq:glo_mv_fea1}) and (\ref{eq:glo_mv_fea2}).
\end{enumerate}

Finally, the total dimension of symmetrized GLO-MV features is $2 \times 2 \times \left( {3 \times 3} \right) = 36$ (the symmetrized GLO-MV is also shown in Table XIII and Fig. 9 in Supplementary).

By analogy, the feature symmetrization process for the GLO-PMV is performed as follows:

\begin{enumerate}[ 1)]
\item Divide the MV set $\Omega ({V_l})$ into \textbf{three} subsets ${\cal V}_l^1 = \left\{ {V_l^1,V_l^3,V_l^7,V_l^9} \right\}$, ${\cal V}_l^2 = \left\{ {V_l^2,V_l^4,V_l^6,V_l^8} \right\}$ and ${\cal V}_l^3 = \left\{ {V_l^5} \right\}$, and all MVs in each subset are regarded as the same MV.

\item For each MV in ${\cal V}_l^1$, divide the PMV set $\Omega ({P_l})$ into \textbf{six} subsets. Taking $V_l^1$ in ${\cal V}_l^1$ as an example, the $\Omega ({P_l})$ is divided into ${\cal P}_l^{1,1} = \left\{ {P_l^1} \right\}$, ${\cal P}_l^{1,2} = \left\{ {P_l^2,P_l^4} \right\}$, ${\cal P}_l^{1,3} = \left\{ {P_l^5} \right\}$, ${\cal P}_l^{1,4} = \left\{ {P_l^3,P_l^7} \right\}$, ${\cal P}_l^{1,5} = \left\{ {P_l^6,P_l^8} \right\}$ and ${\cal P}_l^{1,6} = \left\{ {P_l^9} \right\}$. The set division of $\Omega ({P_l})$ for $V_l^3$, $V_l^7$ and $V_l^9$ is also similar, and the resulted subsets are denoted by ${\cal P}_l^{p,1}, \cdots ,{\cal P}_l^{p,6}$, where $p = 3,7,9$. Merge the subsets ${\cal P}_l^{1,1}$, ${\cal P}_l^{3,1}$, ${\cal P}_l^{7,1}$ and ${\cal P}_l^{9,1}$, and obtain ${\cal P}_l^1 = \mathop  \cup \limits_{p = 1,3,7,9} {\cal P}_l^{p,1}$, where $\cup$ denotes set union. The ${\cal P}_l^2, \cdots ,{\cal P}_l^6$ are also obtained in a similar way. The PMVs in each merged subset are regarded as the same PMV. Combine MV subset ${\cal V}_l^1$ and each PMV subset in ${\cal P}_l^1, \cdots ,{\cal P}_l^6$, and extract the features based on (\ref{eq:glo_pmv_fea1}) and (\ref{eq:glo_pmv_fea2}).

\item For each MV in ${\cal V}_l^2$, divide the PMV set $\Omega ({P_l})$ into \textbf{five} subsets. Taking $V_l^2$ in ${\cal V}_l^2$ as an example, the $\Omega ({P_l})$ is divided into ${\cal P}_l^{2,1} = \left\{ {P_l^2} \right\}$, ${\cal P}_l^{2,2} = \left\{ {P_l^1,P_l^3,P_l^5} \right\}$, ${\cal P}_l^{2,3} = \left\{ {P_l^4,P_l^6} \right\}$, ${\cal P}_l^{2,4} = \left\{ {P_l^8} \right\}$ and ${\cal P}_l^{2,5} = \left\{ {P_l^7,P_l^9} \right\}$. The set division of $\Omega ({P_l})$ for $V_l^4$, $V_l^6$ and $V_l^8$ is also similar, and the resulted subsets are denoted by ${\cal P}_l^{p,1}, \cdots ,{\cal P}_l^{p,5}$, where $p = 4,6,8$. Merge the subsets ${\cal P}_l^{2,1}$, ${\cal P}_l^{4,1}$, ${\cal P}_l^{6,1}$ and ${\cal P}_l^{8,1}$, and obtain ${\cal P}_l^1 = \mathop  \cup \limits_{p = 2,4,6,8} {\cal P}_l^{p,1}$. The ${\cal P}_l^2, \cdots ,{\cal P}_l^5$ are also obtained in a similar way. The PMVs in each merged subset are regarded as the same PMV. Combine MV subset ${\cal V}_l^2$ and each PMV subset in ${\cal P}_l^1, \cdots ,{\cal P}_l^5$, and extract the features based on (\ref{eq:glo_pmv_fea1}) and (\ref{eq:glo_pmv_fea2}).

\item For the MV $V_l^5$ in ${\cal V}_l^3$, divide the PMV set $\Omega ({P_l})$ into \textbf{three} subsets ${\cal P}_l^1 = \left\{ {P_l^1,P_l^3,P_l^7,P_l^9} \right\}$, ${\cal P}_l^2 = \left\{ {P_l^2,P_l^4,P_l^6,P_l^8} \right\}$ and ${\cal P}_l^3 = \left\{ {P_l^5} \right\}$, and the PMVs in each subset are regarded as the same PMV. Combine MV subset ${\cal V}_l^3$ and each PMV subset in ${\cal P}_l^1$, ${\cal P}_l^2$ and ${\cal P}_l^3$ and extract the features based on (\ref{eq:glo_pmv_fea1}) and (\ref{eq:glo_pmv_fea2}).

\item Concatenate the features extracted in setp 2) -- 4) into a feature vector.
\end{enumerate}

%

Finally, the total dimension of symmetrized GLO-PMV features is $2 \times \left( {6{\rm{ + }}5{\rm{ + }}3} \right) = 28$ (the symmetrized GLO-MV is also shown in Table XIV and Fig. 9 in Supplementary).

\section{Experiments} \label{sec:exp}
In this section, we perform extensive experiments to evaluate the effectiveness of the proposed features. We first introduce the experimental settings, and then analyze the feature subsets and validate the feature symmetrization rules. Next, we compare our features with the state-of-the-art methods, and also evaluate the feature robustness under different conditions. Finally, we discuss the computational complexity of features.

\vspace{-2pt}
\subsection{Experimental Setup} \label{sec:exp_setup}
\subsubsection{Video Database}
We use three video databases for experiments, and the databases are introduced as follows.

\textbf{DB1:} this database contains 70 standard test video sequences collected from the internet, as did in \cite{Wang2014AoSO, Zhang2017NPE}, and each of them has 240 frames with a CIF resolution (352$\times$288).

\textbf{DB2:} this database contains 1200 video sequences captured by our digital cameras. The resolution of all the video sequences is resized to 352$\times$288, and the frame length is 36.

\textbf{DB3:} this database is similar to the DB2, but with a resolution of 1920$\times$1080.

The video sequences in three databases are all stored in raw YUV 4:2:0 format with a frame rate of 30 fps.

\begin{table*}[t]
  \newcommand{\tabincell}[2]{\begin{tabular}{@{}#1@{}}#2\end{tabular}}
  \centering
  \begin{minipage}{\columnwidth}
    \caption{Accuracy rates of GLO-MV with different rate-distortion costs}
    \vspace{-5pt}
    \label{tab:steg_glo_mv_sub}
    \centering
    \begin{tabular}{p{31pt}<{\centering}ccccc}
    \toprule
    \multirow{2}{*}[-2pt]{\tabincell{c}{Embedding\\Method}} & \multirow{2}{*}[-2pt]{\tabincell{c}{Feature\\set}} & \multicolumn{2}{c}{QP 15} & \multicolumn{2}{c}{QP 25} \\
    \cmidrule(r){3-4} \cmidrule(l){5-6}
    & & 0.1 & 0.4 & 0.1 & 0.4 \\
    \midrule
    \multirow{3}{*}{Tar1} &
      GLO-MV-SAD-162  & 0.6428 & 0.9161 & 0.6197 & 0.7986 \\
    & GLO-MV-SATD-162 & 0.7182 & 0.8975 & 0.6509 & 0.8262 \\
    & GLO-MV-324      & \textBF{0.7227} & \textBF{0.9178} & \textBF{0.6566} & \textBF{0.8332} \\
    \midrule
    \multirow{3}{*}{Tar2} &
      GLO-MV-SAD-162  & 0.6577 & 0.9089 & 0.6096 & 0.7933 \\
    & GLO-MV-SATD-162 & 0.6667 & 0.8758 & 0.6145 & 0.8196 \\
    & GLO-MV-324      & \textBF{0.6784} & \textBF{0.9166} & \textBF{0.6309} & \textBF{0.8220} \\
    \midrule
    \multirow{3}{*}{Tar3} &
      GLO-MV-SAD-162  & 0.5976 & 0.8741 & 0.6259 & 0.7504 \\
    & GLO-MV-SATD-162 & 0.6377 & 0.8355 & 0.6428 & 0.8019 \\
    & GLO-MV-324      & \textBF{0.6390} & \textBF{0.8934} & \textBF{0.6452} & \textBF{0.8092} \\
    \midrule
    \multirow{3}{*}{Tar4} &
      GLO-MV-SAD-162  & 0.6345 & 0.8352 & 0.5719 & 0.7155 \\
    & GLO-MV-SATD-162 & 0.7512 & 0.8937 & 0.6075 & 0.7647 \\
    & GLO-MV-324      & \textBF{0.7530} & \textBF{0.8947} & \textBF{0.6128} & \textBF{0.7796} \\
    \bottomrule
    \end{tabular}
  \end{minipage}
  \hfill
  \begin{minipage}{\columnwidth}
    \caption{Accuracy rates of GLO-PMV with different rate-distortion costs}
    \vspace{-5pt}
    \label{tab:steg_glo_pmv_sub}
    \centering
    \begin{tabular}{p{31pt}<{\centering}ccccc}
    \toprule
    \multirow{2}{*}[-2pt]{\tabincell{c}{Embedding\\Method}} & \multirow{2}{*}[-2pt]{\tabincell{c}{Feature\\set}} & \multicolumn{2}{c}{QP 15} & \multicolumn{2}{c}{QP 25} \\
    \cmidrule(r){3-4} \cmidrule(l){5-6}
    & & 0.1 & 0.4 & 0.1 & 0.4 \\
    \midrule
    \multirow{3}{*}{Tar1} &
      GLO-PMV-SAD-162  & \textBF{0.6036} & 0.7996 & \textBF{0.6310} & 0.8301 \\
    & GLO-PMV-SATD-162 & 0.5918 & 0.7766 & 0.6287 & 0.8184 \\
    & GLO-PMV-243      & 0.6022 & \textBF{0.8047} & 0.6306 & \textBF{0.8365} \\
    \midrule
    \multirow{3}{*}{Tar2} &
      GLO-PMV-SAD-162  & \textBF{0.5815} & \textBF{0.7989} & 0.6415 & \textBF{0.8226} \\
    & GLO-PMV-SATD-162 & 0.5741 & 0.7767 & \textBF{0.6426} & 0.8191 \\
    & GLO-PMV-243      & 0.5802 & 0.7790 & 0.6352 & 0.8219 \\
    \midrule
    \multirow{3}{*}{Tar3} &
      GLO-PMV-SAD-162  & \textBF{0.5749} & \textBF{0.7557} & 0.6123 & 0.7233 \\
    & GLO-PMV-SATD-162 & 0.5700 & 0.7324 & \textBF{0.6245} & \textBF{0.7304} \\
    & GLO-PMV-243      & 0.5718 & 0.7398 & 0.6156 & 0.7143 \\
    \midrule
    \multirow{3}{*}{Tar4} &
      GLO-PMV-SAD-162  & 0.6046 & \textBF{0.8042} & \textBF{0.6180} & \textBF{0.7541} \\
    & GLO-PMV-SATD-162 & 0.6055 & 0.7861 & 0.6112 & 0.7536 \\
    & GLO-PMV-243      & \textBF{0.6086} & 0.7798 & 0.6103 & 0.7551 \\
    \bottomrule
    \end{tabular}
  \end{minipage}
  \vspace{-5pt}
\end{table*}

\begin{table*}[t]
  \newcommand{\tabincell}[2]{\begin{tabular}{@{}#1@{}}#2\end{tabular}}
  \centering
  \begin{minipage}{\columnwidth}
    \caption{Accuracy rates of GLO-MV before and after feature symmetrization}
    \vspace{-5pt}
    \label{tab:steg_glo_mv_sym}
    \centering
    \begin{tabular}{cccccc}
    \toprule
    \multirow{2}{*}[-2pt]{\tabincell{c}{Embedding\\Method}} & \multirow{2}{*}[-2pt]{\tabincell{c}{Feature\\set}} & \multicolumn{2}{c}{QP 15} & \multicolumn{2}{c}{QP 25} \\
    \cmidrule(r){3-4} \cmidrule(l){5-6}
    & & 0.1 & 0.4 & 0.1 & 0.4 \\
    \midrule
    \multirow{2}{*}{Tar1} &
      GLO-MV-324 & 0.7227 & 0.9178 & \textBF{0.6566} & 0.8332 \\
    & GLO-MV-36  & \textBF{0.7299} & \textBF{0.9207} & 0.6548 & \textBF{0.8462} \\
    \midrule
    \multirow{2}{*}{Tar2} &
      GLO-MV-324 & 0.6784 & 0.9166 & \textBF{0.6309} & 0.8220 \\
    & GLO-MV-36  & \textBF{0.6948} & \textBF{0.9207} & 0.6229 & \textBF{0.8302} \\
    \midrule
    \multirow{2}{*}{Tar3} &
      GLO-MV-324 & 0.6390 & \textBF{0.8934} & \textBF{0.6452} & 0.8092 \\
    & GLO-MV-36  & \textBF{0.6485} & 0.8715 & 0.6368 & \textBF{0.8124} \\
    \midrule
    \multirow{2}{*}{Tar4} &
      GLO-MV-324 & \textBF{0.7530} & 0.8947 & \textBF{0.6128} & \textBF{0.7796} \\
    & GLO-MV-36  & 0.7327 & \textBF{0.8956} & 0.6103 & 0.7671 \\
    \bottomrule
    \end{tabular}
  \end{minipage}
  \hfill
  \begin{minipage}{\columnwidth}
    \caption{Accuracy rates of GLO-PMV before and after feature symmetrization}
    \vspace{-5pt}
    \label{tab:steg_glo_pmv_sym}
    \centering
    \begin{tabular}{cccccc}
    \toprule
    \multirow{2}{*}[-2pt]{\tabincell{c}{Embedding\\Method}} & \multirow{2}{*}[-2pt]{\tabincell{c}{Feature\\set}} & \multicolumn{2}{c}{QP 15} & \multicolumn{2}{c}{QP 25} \\
    \cmidrule(r){3-4} \cmidrule(l){5-6}
    & & 0.1 & 0.4 & 0.1 & 0.4 \\
    \midrule
    \multirow{2}{*}{Tar1} &
      GLO-PMV-162 & 0.6036 & 0.7996 & 0.6310 & 0.8301 \\
    & GLO-PMV-28  & \textBF{0.6303} & \textBF{0.8780} & \textBF{0.6517} & \textBF{0.8624} \\
    \midrule
    \multirow{2}{*}{Tar2} &
      GLO-PMV-162 & 0.5815 & 0.7989 & 0.6415 & 0.8226 \\
    & GLO-PMV-28  & \textBF{0.6026} & \textBF{0.8797} & \textBF{0.6451} & \textBF{0.8523} \\
    \midrule
    \multirow{2}{*}{Tar3} &
      GLO-PMV-162 & 0.5749 & 0.7557 & 0.6123 & 0.7233 \\
    & GLO-PMV-28  & \textBF{0.5846} & \textBF{0.7827} & \textBF{0.6381} & \textBF{0.7547} \\
    \midrule
    \multirow{2}{*}{Tar4} &
      GLO-PMV-162 & 0.6046 & \textBF{0.8042} & 0.6180 & 0.7541 \\
    & GLO-PMV-28  & \textBF{0.6074} & 0.7731 & \textBF{0.6216} & \textBF{0.7757} \\
    \bottomrule
    \end{tabular}
  \end{minipage}
  \vspace{-5pt}
\end{table*}

\begin{table}[t]
  \newcommand{\tabincell}[2]{\begin{tabular}{@{}#1@{}}#2\end{tabular}}
  \caption{Accuracy rates of GLO before and after feature symmetrization}
  \vspace{-5pt}
  \label{tab:steg_glo_sym}
  \centering
  \begin{tabular}{cccccc}
    \toprule
    \multirow{2}{*}[-2pt]{\tabincell{c}{Embedding\\Method}} & \multirow{2}{*}[-2pt]{\tabincell{c}{Feature\\set}} & \multicolumn{2}{c}{QP 15} & \multicolumn{2}{c}{QP 25} \\
    \cmidrule(r){3-4} \cmidrule(l){5-6}
    & & 0.1 & 0.4 & 0.1 & 0.4 \\
    \midrule
    \multirow{2}{*}{Tar1} &
      GLO-486 & 0.7242 & 0.9373 & 0.6845 & 0.8697 \\
    & GLO-64  & \textBF{0.7332} & \textBF{0.9475} & \textBF{0.6978} & \textBF{0.9011} \\
    \midrule
    \multirow{2}{*}{Tar2} &
      GLO-486 & 0.6789 & 0.9228 & 0.6652 & 0.8454 \\
    & GLO-64  & \textBF{0.7035} & \textBF{0.9393} & \textBF{0.6838} & \textBF{0.8769} \\
    \midrule
    \multirow{2}{*}{Tar3} &
      GLO-486 & 0.6432 & 0.9059 & 0.6673 & \textBF{0.8238} \\
    & GLO-64  & \textBF{0.6577} & \textBF{0.9200} & \textBF{0.6884} & 0.8210 \\
    \midrule
    \multirow{2}{*}{Tar4} &
      GLO-486 & \textBF{0.7565} & 0.9012 & \textBF{0.6576} & 0.8026 \\
    & GLO-64  & 0.7415 & \textBF{0.9018} & 0.6475 & \textBF{0.8130} \\
    \bottomrule
  \end{tabular}
  \vspace{-7pt}
\end{table}

\subsubsection{Steganographic Methods}
We use four steganographic methods to evaluate the detection performance of our features, including Cao \etal's method \cite{Cao2011PME} (statistics-preserved, denoted as Tar1), Cao \etal's method \cite{Cao2015PMEopt} (local optimality-preserved, denoted as Tar2), Zhang \etal's method \cite{Zhang2015Preserved} (local optimality-preserved, denoted as Tar3) and Zhu \etal's method \cite{Zhu2018UED} (combination of statistics and local optimality, denoted as Tar4). The embedding rate is measured by bpnsmv (bits per non-skip motion vector), which is the ratio of the number of embedded bits to the total number of non-skip MVs. Unless stated otherwise, all the steganographic methods are implemented in a H.264/AVC encoder x264 \cite{x264} with constant quantization parameter (CQP), and the block matching method is Hexagon Search algorithm (HEX).

\subsubsection{Steganalytic Methods}
We compare our method with two state-of-the-art steganalytic methods based on local optimality \ie, AoSO \cite{Wang2014AoSO} and NPE \cite{Zhang2017NPE}. We also use the universal steganalytic method \cite{Zhai2019MVC} and recent HEVC-specified steganalytic methods \cite{Liu2021LOCL} for comparison, but we provide their resuts in Supplementary due to limited space.

\subsubsection{Training and Testing}
We implement steganalyzers using a Gaussian-kernel SVM (support vector machine) \cite{libsvm}, whose hyper-parameters $\left( {C,\gamma } \right)$ are optimized by a grid search on  $\left\{ {\left. {\left( {C,\gamma } \right)} \;\right|C = {2^{ - 5}},{2^{ - 4}}, \cdots ,{2^{15}},\;\gamma  = {2^{ - 15}},{2^{ - 14}}, \cdots ,{2^3}} \right\}$. The cover and stego pairs of the video sequences in each database are randomly split into two equal-sized halves as a training set and a testing set. The detection performance is measured by accuracy rate, which is the mean value of true positive rate and true negative rate, and the final accuracy rate is averaged over 20 random splits of the database.


\begin{figure*}[t]
  \centering
  \subfloat{\includegraphics[width=0.25\textwidth]{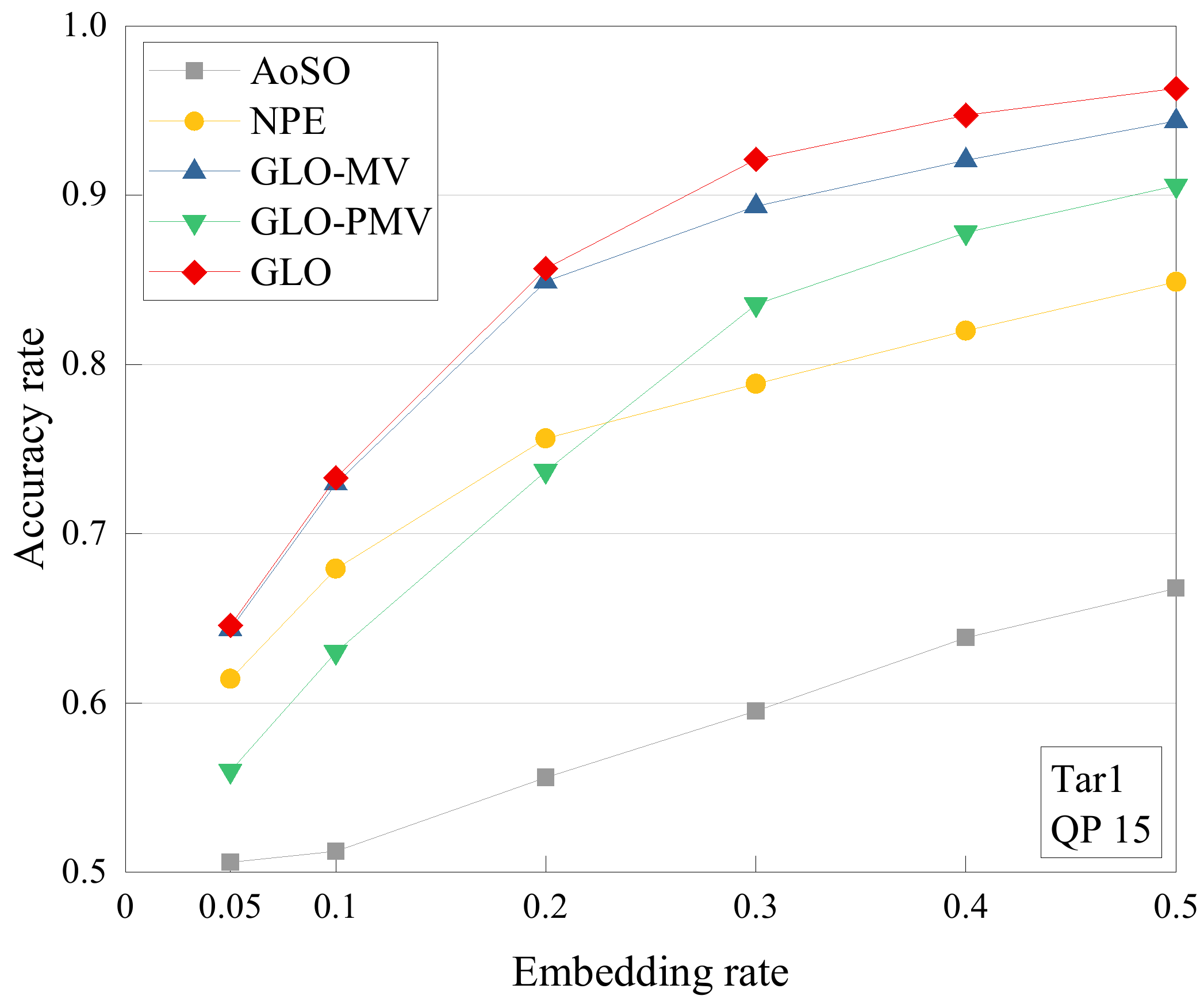}\label{fig:steg_glo_tar1_qp15}}
  \hfil
  \subfloat{\includegraphics[width=0.25\textwidth]{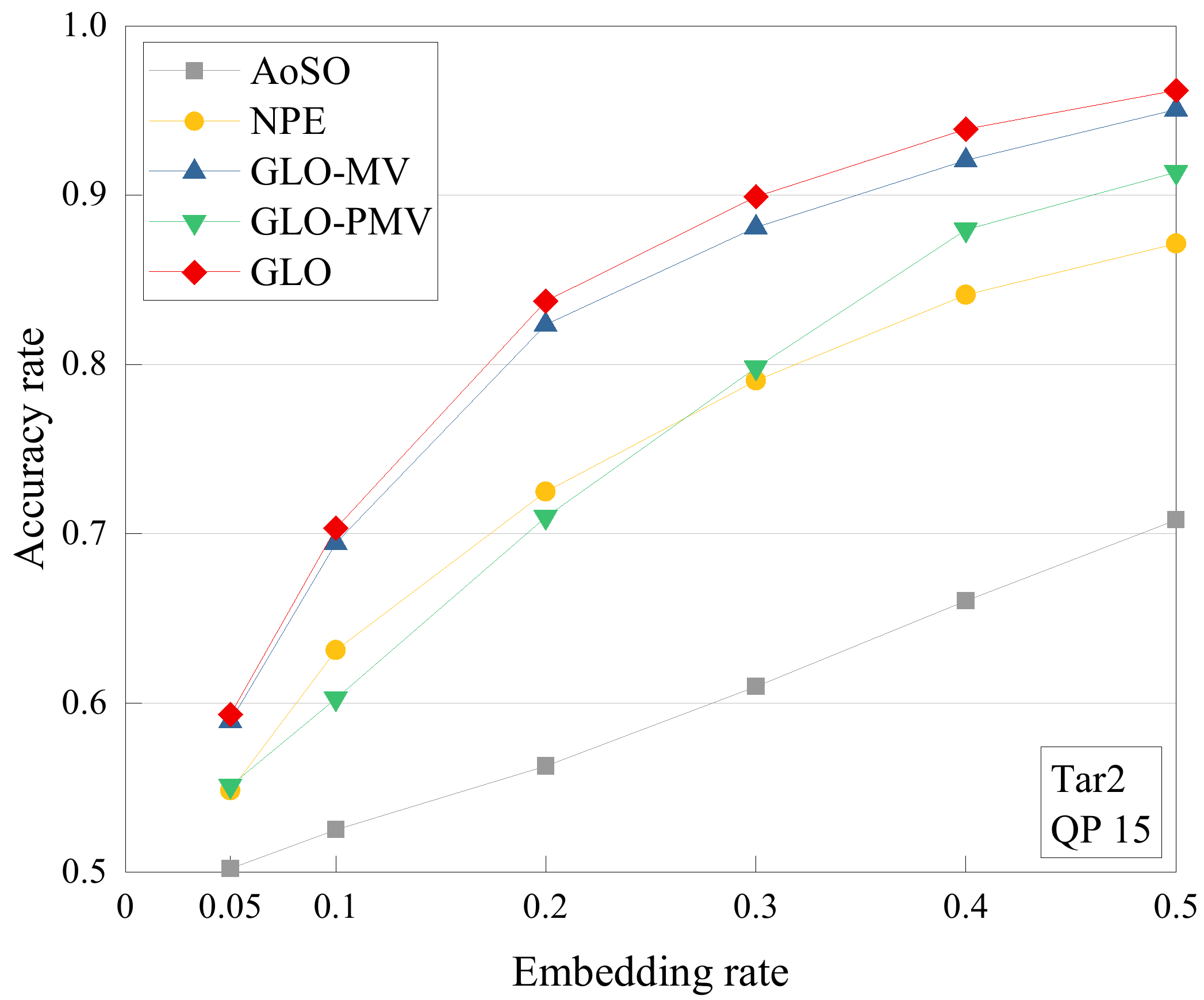}\label{fig:steg_glo_tar2_qp15}}
  \hfil
  \subfloat{\includegraphics[width=0.25\textwidth]{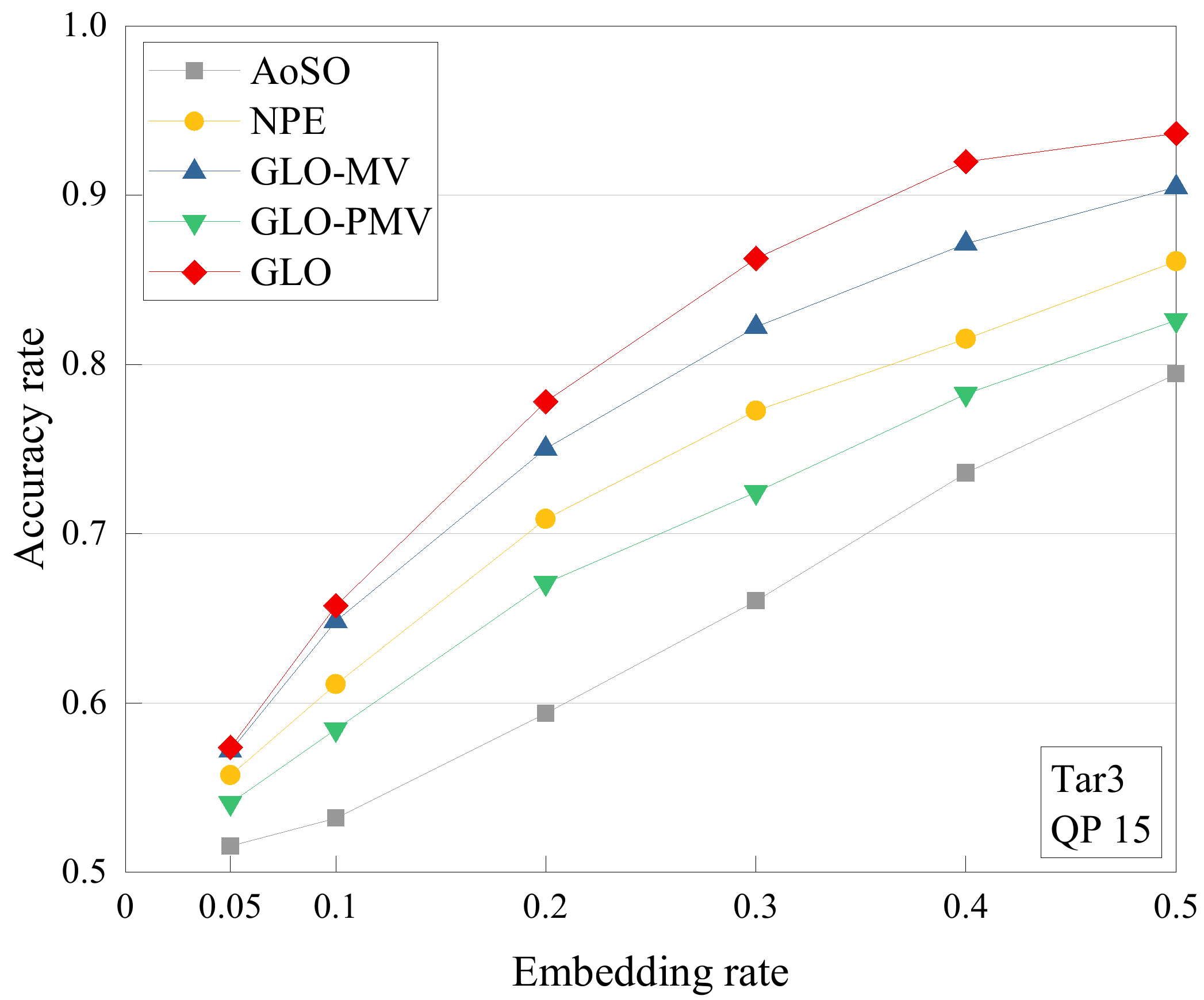}\label{fig:steg_glo_tar3_qp15}}
  \hfil
  \subfloat{\includegraphics[width=0.25\textwidth]{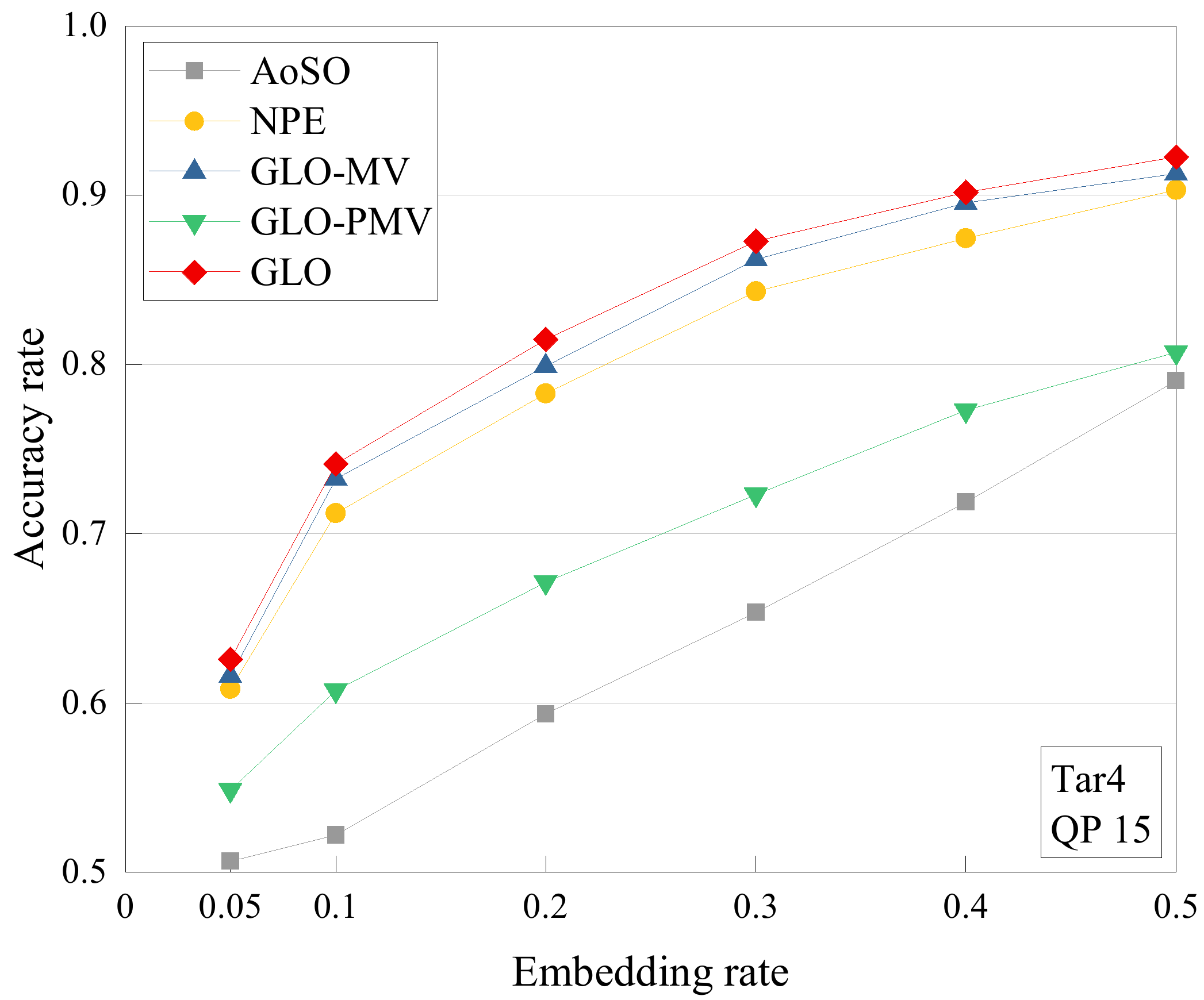}\label{fig:steg_glo_tar4_qp15}}
  \vfil
  \vspace{-10pt}
  \subfloat{\includegraphics[width=0.25\textwidth]{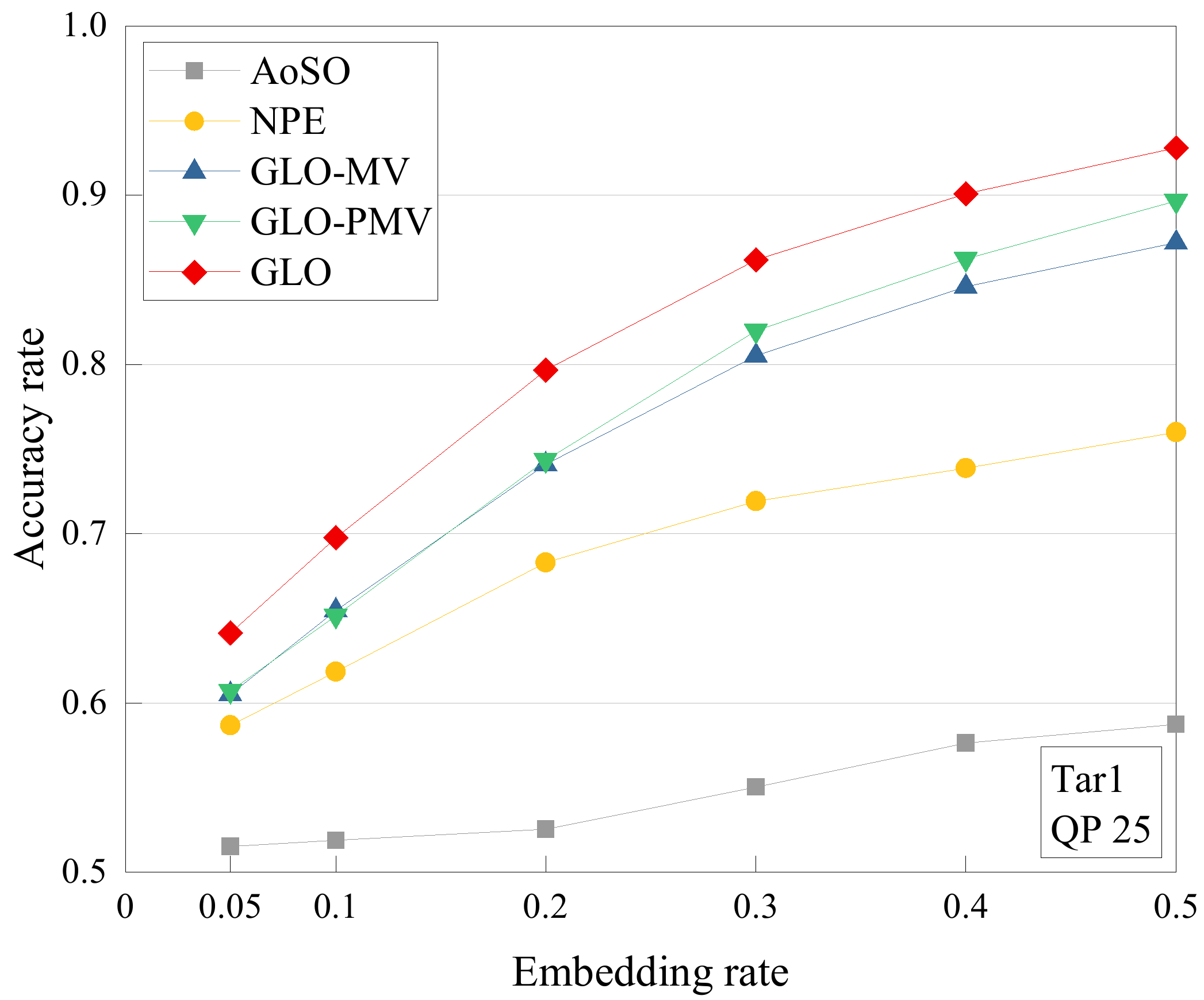}\label{fig:steg_glo_tar1_qp25}}
  \hfil
  \subfloat{\includegraphics[width=0.25\textwidth]{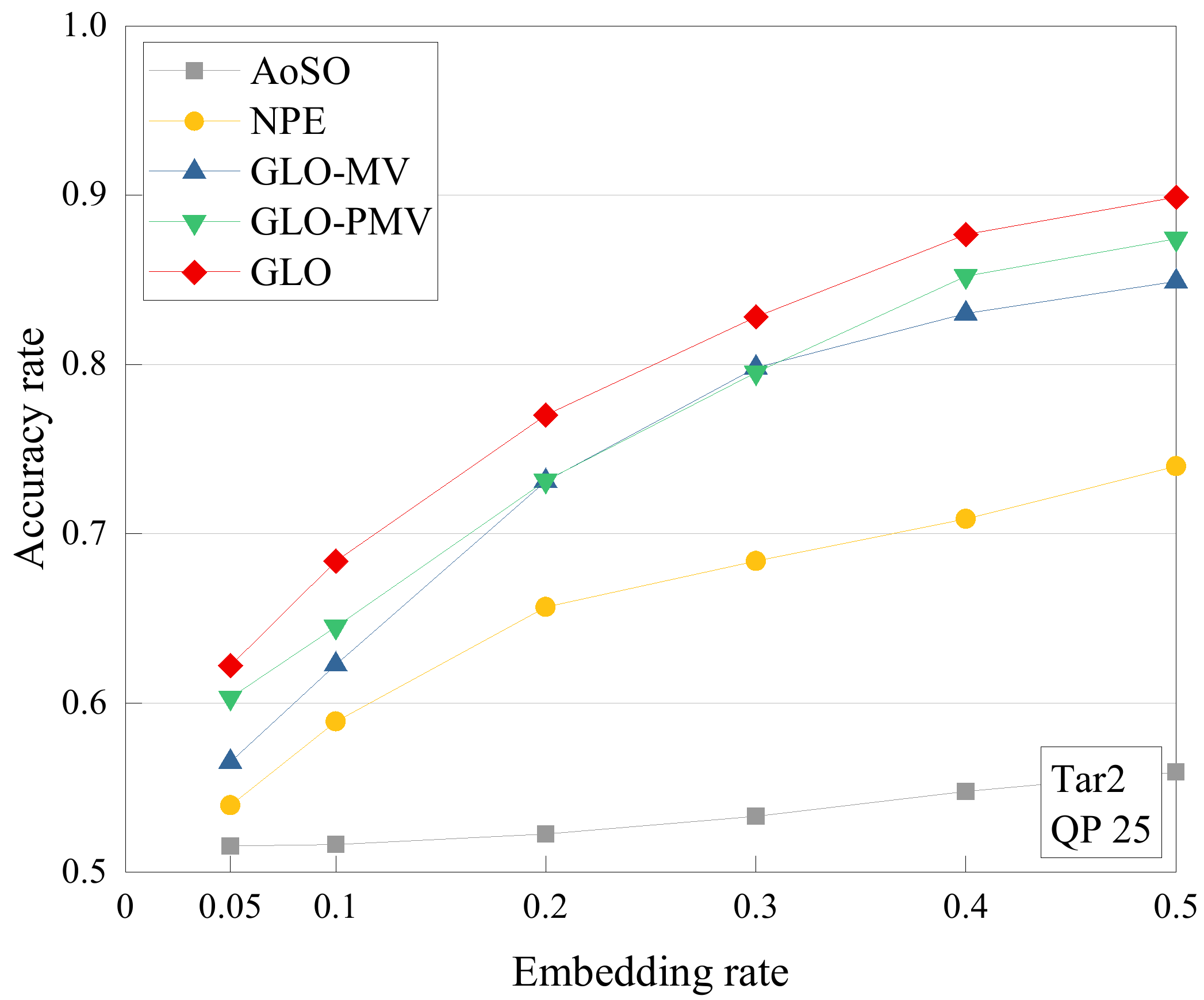}\label{fig:steg_glo_tar2_qp25}}
  \hfil
  \subfloat{\includegraphics[width=0.25\textwidth]{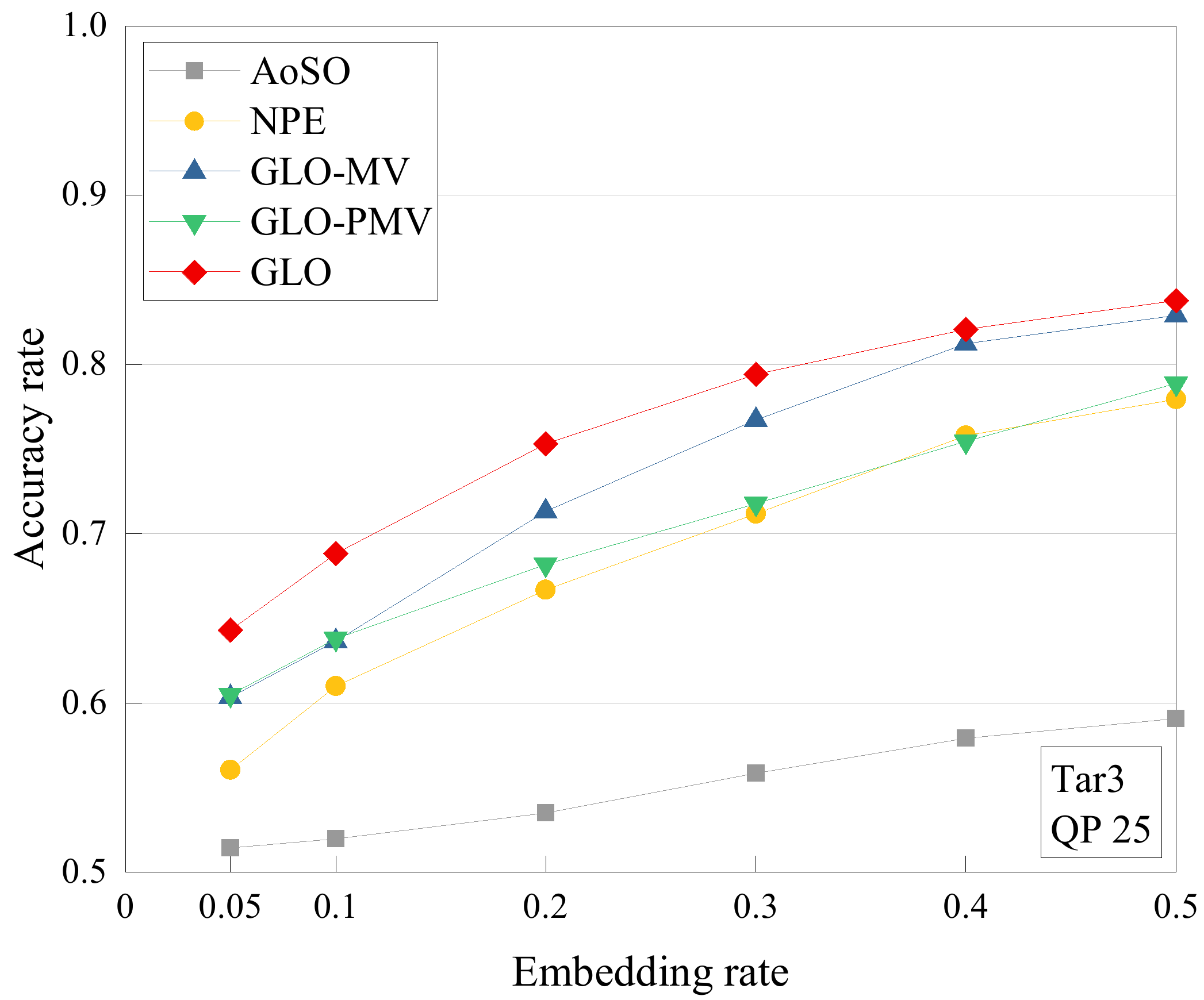}\label{fig:steg_glo_tar3_qp25}}
  \hfil
  \subfloat{\includegraphics[width=0.25\textwidth]{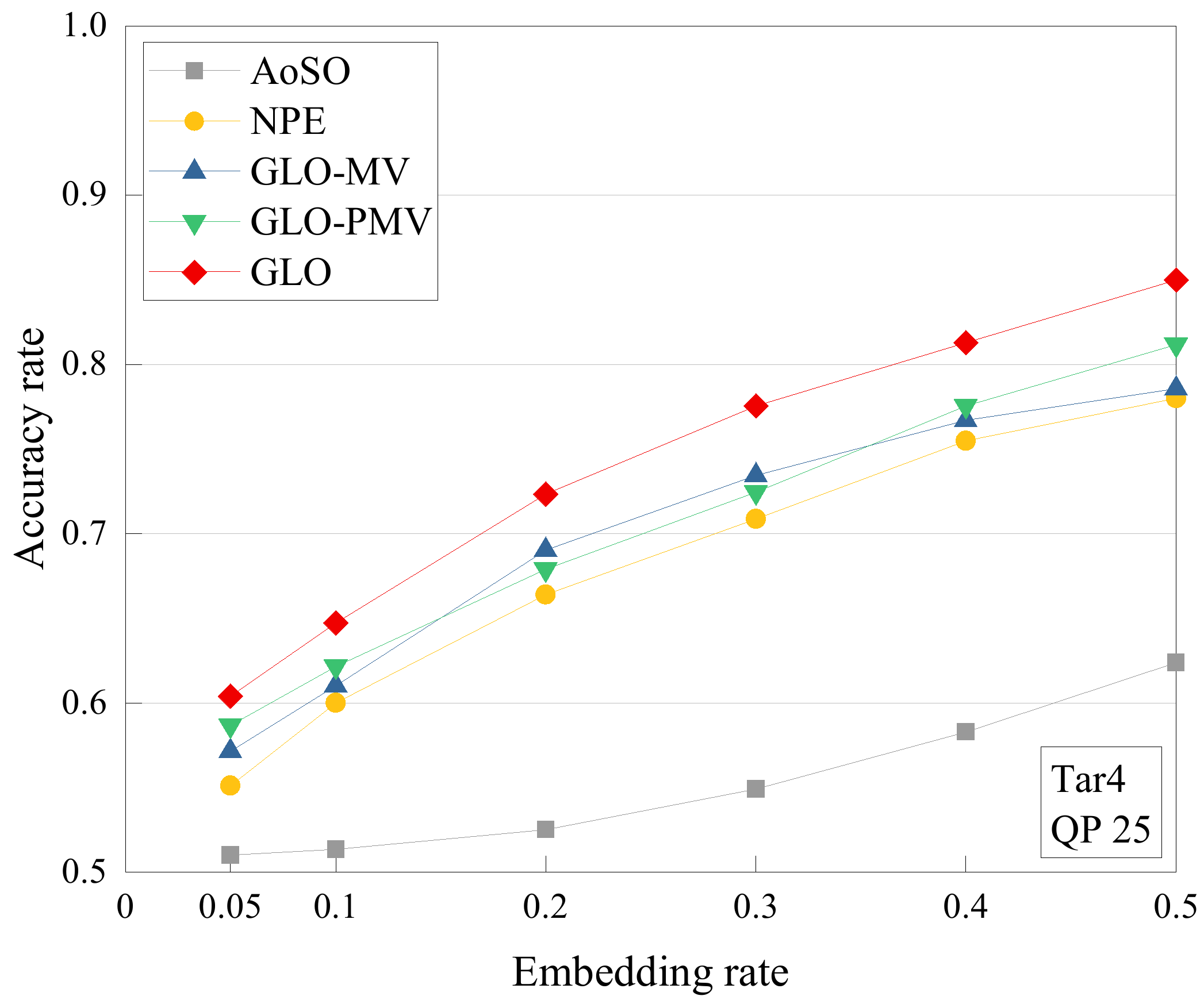}\label{fig:steg_glo_tar4_qp25}}
  \vspace{-8pt}
  \caption{Accuracy rates of AoSO \cite{Wang2014AoSO}, NPE \cite{Zhang2017NPE}, and GLO and its subfeatures as a function of embedding rate for four MV based steganographic methods on DB1 with QP 15 and 25.}
  \vspace{-6pt}
  \label{fig:steg_glo}
\end{figure*}

\subsection{Feature Subset Analysis} \label{sec:exp_subfea}
We adopt two types of rate-distortion costs to measure the generalized local optimality. To evaluate the effect of SAD and SATD based rate-distortion costs on the detection performance of the proposed features, we report the accuracy rates of GLO-MV and GLO-PMV with different rate-distortion costs in Table \ref{tab:steg_glo_mv_sub} and Table \ref{tab:steg_glo_pmv_sub}, where the best result is marked in bold for each group of comparison. The suffixes ``SAD'' and ``SATD'' in the steganalytic method names mean SAD and SATD based rate-distortion costs, and the number suffixes indicate feature dimensions. The experiments are conducted on DB1 with QP 15 and 25, and the steganographic methods are Tar1–4 with 0.1 and 0.4 bpnsmv.

It can be seen from Table \ref{tab:steg_glo_mv_sub} that the SATD based rate-distortion cost contributes more to the GLO-MV in most cases, while the SAD based rate-distortion cost becomes more important at high payloads and low QP values. This indicates that the SAD and SATD based rate-distortion costs play different roles in different conditions. Besides, combining the two types of rate-distortion costs always has better detection performance than using one of them alone, demonstrating that the SAD and SATD based rate-distortion costs should be both considered for the GLO-MV.

As seen from Table \ref{tab:steg_glo_pmv_sub}, the accuracy rates of GLO-PMV-SAD-162 and GLO-PMV-SATD-162 are basically the same, even combining them does not necessarily improve the detection performance. Because the GLO-PMV depends mainly on the changes of PMVs, and the prediction error (\ie, distortion) remains unchanged when estimating the generalized local optimality of PMVs. As a result, both SAD and SATD have roughly the same impact on the GLO-PMV, and that is why we only use SAD based rate-distortion cost in (\ref{eq:glo_pmv_fea1}) and (\ref{eq:glo_pmv_fea2}).

\subsection{Feature Symmetrization Validation} \label{sec:exp_symfea}

To validate the effectiveness of  feature symmetrization rules, we compare the detection performance of GLO-MV and GLO-PMV before and after feature symmetrization, and list their results in Table \ref{tab:steg_glo_mv_sym} and Table \ref{tab:steg_glo_pmv_sym}. In each comparison, the larger suffix number in the method name represents the dimension before feature symmetrization, and the smaller suffix number represents the dimension after feature symmetrization. The used video database, QPs and embedding rates are the same as those in the last subsection.

We observe from Table \ref{tab:steg_glo_mv_sym} that the detection performance of GLO-MV before and after feature symmetrization is roughly the same. In many cases, the accuracy rates of symmetrized features are even higher than those of original features. In Table \ref{tab:steg_glo_pmv_sym}, the GLO-PMV-28 performs better than GLO-PMV-162 in nearly all cases. This verifies that there exists redundancy in original features, and the feature symmetrization can reduce feature dimension without compromising detection accuracy.

The GLO-MV and the GLO-PMV reflect the embedding changes from different aspects, so it is necessary to merge these two types of features. We also report in Table \ref{tab:steg_glo_sym} the accurate rates of GLO (the combination of GLO-MV and the GLO-PMV) before and after feature symmetrization. We can see that the symmetrized features (GLO-64) still have better detection performance in nearly all cases, again demonstrating the importance of feature symmetrization. Therefore, we adopt the symmetrized features for subsequent experiments.

\begin{table*}[t]
  \newcommand{\tabincell}[2]{\begin{tabular}{@{}#1@{}}#2\end{tabular}}
  \centering
  \begin{minipage}{\columnwidth}
    \caption{Accuracy rates of NPE \cite{Zhang2017NPE}, and GLO and its subfeatures on DB1 and DB2 in a cover-source mismatch scenario}
    \vspace{-5pt}
    \label{tab:steg_csm}
    \centering
    \begin{tabular}{cccccc}
      \toprule
      \multirow{2}{*}[-2pt]{\tabincell{c}{Embedding\\Method}} & \multirow{2}{*}[-2pt]{\tabincell{c}{Feature\\Set}} & \multicolumn{2}{c}{QP 15} & \multicolumn{2}{c}{QP 25} \\
      \cmidrule(r){3-4} \cmidrule(l){5-6}
      & & 0.1 & 0.4 & 0.1 & 0.4 \\
      \midrule
      \multirow{4}{*}{Tar1} &
        NPE     & 0.7404 & 0.8537 & 0.6445 & 0.7644 \\
      & GLO-MV  & 0.7833 & 0.9485 & 0.6525 & 0.8427 \\
      & GLO-PMV & 0.6480 & 0.8556 & 0.6420 & 0.8580 \\
      & GLO     & \textBF{0.7925} & \textBF{0.9564} & \textBF{0.6775} & \textBF{0.8754} \\
      \midrule
      \multirow{4}{*}{Tar2} &
        NPE     & 0.6901 & 0.8796 & 0.5756 & 0.7270 \\
      & GLO-MV  & 0.7289 & 0.9425 & 0.6177 & 0.8347 \\
      & GLO-PMV & 0.6115 & 0.8795 & 0.6400 & 0.8448 \\
      & GLO     & \textBF{0.7364} & \textBF{0.9522} & \textBF{0.6520} & \textBF{0.8674} \\
      \midrule
      \multirow{4}{*}{Tar3} &
        NPE     & 0.6379 & 0.8597 & 0.5956 & 0.7532 \\
      & GLO-MV  & 0.6698 & 0.9054 & 0.6406 & 0.8017 \\
      & GLO-PMV & 0.5810 & 0.7742 & 0.5985 & 0.7302 \\
      & GLO     & \textBF{0.6758} & \textBF{0.9283} & \textBF{0.6470} & \textBF{0.8184} \\
      \midrule
      \multirow{4}{*}{Tar4} &
        NPE     & 0.7890 & 0.9061 & 0.6075 & 0.7508 \\
      & GLO-MV  & 0.7990 & 0.9398 & 0.6112 & 0.7646 \\
      & GLO-PMV & 0.5991 & 0.7491 & 0.6084 & 0.7672 \\
      & GLO     & \textBF{0.8052} & \textBF{0.9402} & \textBF{0.6356} & \textBF{0.7711} \\
      \bottomrule
    \end{tabular}
  \end{minipage}
  \hfill
  \begin{minipage}{\columnwidth}
    \caption{Accuracy rates of NPE \cite{Zhang2017NPE}, and GLO and its subfeatures on DB1 using different block matching algorithms}
    \vspace{-5pt}
    \label{tab:steg_bma}
    \centering
    \begin{tabular}{cccccc}
      \toprule
      \multirow{2}{*}[-2pt]{\tabincell{c}{Embedding\\Method}} & \multirow{2}{*}[-2pt]{\tabincell{c}{Feature\\Set}} & \multicolumn{2}{c}{QP 15} & \multicolumn{2}{c}{QP 25} \\
      \cmidrule(r){3-4} \cmidrule(l){5-6}
      & & 0.1 & 0.4 & 0.1 & 0.4 \\
      \midrule
      \multirow{4}{*}{\tabincell{c}{Tar2\\(DIA)}} &
        NPE     & 0.6465 & 0.8161 & 0.5859 & 0.6975 \\
      & GLO-MV  & 0.6754 & 0.9225 & 0.6285 & 0.8292 \\
      & GLO-PMV & 0.6686 & 0.8652 & 0.6614 & 0.8413 \\
      & GLO     & \textBF{0.7088} & \textBF{0.9316} & \textBF{0.6834} & \textBF{0.8624} \\
      \midrule
      \multirow{4}{*}{\tabincell{c}{Tar2\\(HEX)}} &
        NPE     & 0.6312 & 0.8413 & 0.5891 & 0.7089 \\
      & GLO-MV  & 0.6948 & 0.9207 & 0.6229 & 0.8302 \\
      & GLO-PMV & 0.6026 & 0.8797 & 0.6451 & 0.8523 \\
      & GLO     & \textBF{0.7035} & \textBF{0.9393} & \textBF{0.6838} & \textBF{0.8769} \\
      \midrule
      \multirow{4}{*}{\tabincell{c}{Tar2\\(UMH)}} &
        NPE     & 0.6336 & 0.8423 & 0.5818 & 0.7040 \\
      & GLO-MV  & 0.6994 & 0.9218 & 0.6369 & 0.8444 \\
      & GLO-PMV & 0.6270 & 0.8726 & 0.6634 & 0.8562 \\
      & GLO     & \textBF{0.7037} & \textBF{0.9384} & \textBF{0.6840} & \textBF{0.8647} \\
      \midrule
      \multirow{4}{*}{\tabincell{c}{Tar2\\(ESA)}} &
        NPE     & 0.6853 & 0.8641 & 0.6435 & 0.7225 \\
      & GLO-MV  & 0.7707 & 0.9309 & 0.6992 & 0.8675 \\
      & GLO-PMV & 0.8007 & 0.9102 & 0.7333 & 0.8692 \\
      & GLO     & \textBF{0.8125} & \textBF{0.9419} & \textBF{0.7632} & \textBF{0.8811} \\
      \bottomrule
    \end{tabular}
  \end{minipage}
  \vfill
  \vspace{-5pt}
  \vspace{10pt}
  \begin{minipage}{\columnwidth}
    \caption{Accuracy rates of NPE \cite{Zhang2017NPE}, and GLO and its subfeatures on DB3 using different video resolutions}
    \vspace{-5pt}
    \label{tab:steg_resolution}
    \centering
    \begin{tabular}{cccccc}
      \toprule
      \multirow{2}{*}[-2pt]{\tabincell{c}{Embedding\\Method}} & \multirow{2}{*}[-2pt]{\tabincell{c}{Feature\\Set}} & \multicolumn{2}{c}{QP 15} & \multicolumn{2}{c}{QP 25} \\
      \cmidrule(r){3-4} \cmidrule(l){5-6}
      & & 0.1 & 0.4 & 0.1 & 0.4 \\
      \midrule
      \multirow{4}{*}{\tabincell{c}{Tar1}} &
        NPE     & 0.7860 & 0.9481 & 0.7442 & 0.8655 \\
      & GLO-MV  & 0.8357 & 0.9915 & 0.7638 & 0.9519 \\
      & GLO-PMV & 0.7635 & 0.9761 & 0.7760 & 0.9690 \\
      & GLO     & \textBF{0.8480} & \textBF{0.9923} & \textBF{0.8155} & \textBF{0.9760} \\
      \midrule
      \multirow{4}{*}{\tabincell{c}{Tar2}} &
        NPE     & 0.7695 & 0.9547 & 0.7165 & 0.8287 \\
      & GLO-MV  & 0.8223 & 0.9890 & 0.7434 & 0.9492 \\
      & GLO-PMV & 0.7459 & 0.9824 & 0.7747 & 0.9633 \\
      & GLO     & \textBF{0.8305} & \textBF{0.9906} & \textBF{0.8042} & \textBF{0.9652} \\
      \midrule
      \multirow{4}{*}{\tabincell{c}{Tar3}} &
        NPE     & 0.7476 & 0.9266 & 0.7354 & 0.8910 \\
      & GLO-MV  & 0.7829 & 0.9728 & 0.7728 & 0.9298 \\
      & GLO-PMV & 0.7114 & 0.9216 & 0.7702 & 0.8967 \\
      & GLO     & \textBF{0.7918} & \textBF{0.9882} & \textBF{0.8063} & \textBF{0.9346} \\
      \midrule
      \multirow{4}{*}{\tabincell{c}{Tar4}} &
        NPE     & 0.8064 & 0.9758 & 0.7268 & 0.8845 \\
      & GLO-MV  & 0.8416 & 0.9829 & 0.7408 & 0.8946 \\
      & GLO-PMV & 0.7365 & 0.9504 & 0.7572 & 0.9094 \\
      & GLO     & \textBF{0.8523} & \textBF{0.9855} & \textBF{0.7875} & \textBF{0.9213} \\
      \bottomrule
    \end{tabular}
  \end{minipage}
  \hfill
  \begin{minipage}{\columnwidth}
    \caption{Accuracy rates of NPE \cite{Zhang2017NPE}, and GLO and its subfeatures on DB1 created by MPEG-4 Visual standard}
    \vspace{-5pt}
    \label{tab:steg_mpeg4}
    \centering
    \begin{tabular}{cccccc}
      \toprule
      \multirow{2}{*}[-2pt]{\tabincell{c}{Embedding\\Method}} & \multirow{2}{*}[-2pt]{\tabincell{c}{Feature\\Set}} & \multicolumn{2}{c}{1000 kb/s} & \multicolumn{2}{c}{500 kb/s} \\
      \cmidrule(r){3-4} \cmidrule(l){5-6}
      & & 0.1 & 0.4 & 0.1 & 0.4 \\
      \midrule
      \multirow{4}{*}{Tar1} &
        NPE     & 0.6008 & 0.7804 & 0.5581 & 0.7043 \\
      & GLO-MV  & 0.6523 & 0.8712 & 0.6010 & 0.7972 \\
      & GLO-PMV & 0.5506 & 0.8187 & 0.5971 & 0.8368 \\
      & GLO     & \textBF{0.6616} & \textBF{0.9185} & \textBF{0.6389} & \textBF{0.9001} \\
      \midrule
      \multirow{4}{*}{Tar2} &
        NPE     & 0.5602 & 0.8124 & 0.5311 & 0.6784 \\
      & GLO-MV  & 0.6247 & 0.8732 & 0.5683 & 0.7937 \\
      & GLO-PMV & 0.5498 & 0.8254 & 0.6281 & 0.8186 \\
      & GLO     & \textBF{0.6275} & \textBF{0.8925} & \textBF{0.6390} & \textBF{0.8508} \\
      \midrule
      \multirow{4}{*}{Tar3} &
        NPE     & 0.5442 & 0.7713 & 0.5498 & 0.7240 \\
      & GLO-MV  & 0.5786 & 0.8322 & 0.5828 & 0.7877 \\
      & GLO-PMV & 0.5312 & 0.7684 & 0.5745 & 0.7038 \\
      & GLO     & \textBF{0.5905} & \textBF{0.8710} & \textBF{0.6257} & \textBF{0.8104} \\
      \midrule
      \multirow{4}{*}{Tar4} &
        NPE     & 0.6474 & 0.8387 & 0.5450 & 0.7254 \\
      & GLO-MV  & 0.6605 & 0.8717 & 0.5561 & 0.7386 \\
      & GLO-PMV & 0.5353 & 0.7046 & 0.5629 & 0.7406 \\
      & GLO     & \textBF{0.6672} & \textBF{0.8785} & \textBF{0.5992} & \textBF{0.7944} \\
      \bottomrule
    \end{tabular}
  \end{minipage}
\end{table*}

\subsection{Comparison With Prior Art} \label{sec:exp_cmp}

To demonstrate the superiority of the proposed features based on generalized local optimality, we compare our method with state-of-the-art methods AoSO and NPE. The GLO-MV, GLO-PMV and GLO are all symmetrized features with dimensions 36-D, 28-D and 64-D, and the AoSO and NPE are 18-D and 36-D features. The experiments are conducted with Tar1–4 on DB1 at QP 15 and 25. The accuracy rates of different steganalytic methods as a function of embedding rate for four steganographic methods at two QPs are shown in Fig. \ref{fig:steg_glo} (the exact numerical values of all accuracy rates are also reported in Table XVI in Supplementary).

For the comparisons between GLO-MV and previous features, we can see that the GLO-MV consistently outperforms the AoSO and NPE by a significant margin in all cases, indicating that the generalized local optimality of MVs is easier to reveal the embedding changes caused by MV based steganography. The feature dimensions of GLO-MV and NPE are the same, but the accuracy rate of GLO-MV is up to 12.13\% higher than that of NPE (see the results of Tar2 in 0.4 bpnsmv at QP 25).

For the GLO-PMV at QP 15, its detection performance is always lower than GLO-MV. By comparison with AoSO and NPE, the GLO-PMV clearly outperforms AoSO and also surpasses NPE for high payloads, but is inferior to NPE for low payloads. The reasons are as follows. First, the rate-distortion cost at a low QP is dominated by the prediction error (PE), and the GLO-PMV pays less attention to the change of PE. Second, the change rate of PMVs is lower than that of MVs, especially in low payloads (see Fig. \ref{fig:PMV_bitrate}). While for the GLO-PMV at QP 25, its detection performance is significantly improved. The accuracy rate of GLO-PMV is basically higher than that of NPE (by a margin as high as 14.34\% against Tar2 in 0.4 bpnsmv at QP 25), and even exceeds GLO-MV when detecting Tar1 and Tar2. This is because high QP leads to greater quantization distortion \cite{Wang2014AoSO}, which affects more on the PE. Compared with AoSO, NPE and GLO-MV that need to calculate the PEs corresponding to various changes of the MVs, the GLO-PMV only needs to calculate the PE of a given MV, so it is less affected by the PE, and is also more robust to the quantization distortion.

For the GLO which combines GLO-MV and GLO-PMV, it provides the best performance across all steganographic methods, embedding rates and QPs, confirming the necessity of feature construction from different aspects of generalized local optimality.

\subsection{Evaluation of Feature Robustness} \label{sec:exp_robust}

\subsubsection{Steganalysis on Mismatched Cover Sources}
The mismatch between training set and testing set severely degrades the steganalysis performance. To evaluate the robustness of the proposed features to the mismatched cover sources, we train the steganalyzers on DB1 and test them on DB2. The DB1 and the DB2 differ greatly in imaging devices and video content, and can well simulate the source mismatch occurred in the real-world. For simplicity, we only use NPE for comparison, and only consider two payloads 0.1 and 0.4 bpnsmv. The accuracy rates of NPE, GLO-MV, GLO-PMV and GLO in a mismatched scenario are reported in Table \ref{tab:steg_csm}.

As seen from Table \ref{tab:steg_csm}, the GLO and its sub-features can still obtain relatively stable detection performance under mismatched sources. Specifically, the accuracy rate of GLO-PMV is generally lower than NPE at a low QP, but higher than NPE at a high QP, which is consistent with that in matched sources in Fig. \ref{fig:steg_glo}. The accuracy rates of GLO-MV and GLO are significantly higher than that of NPE in all conditions, indicating that the generalized local optimality has a good robustness in the mismatched scenario.

\subsubsection{Steganalysis under Different Block Matching Algorithms}
The block matching algorithm used in motion estimation (ME) affects the prediction accuracy of blocks, and also affects the rate-distortion costs. To evaluate the effect of block matching algorithms on the steganalytic features, we compress the video sequences by different algorithms, including Diamond Search (DIA), Hexagonal Search (HEX), Uneven Multi-Hexagon Search (UMH) and Exhaustive Search (ESA). The experiments are performed on DB1 at QP 15 and 25, and only Tar2 is used for simplifying the comparisons. The accuracy rates of NPE, GLO-MV, GLO-PMV and GLO under different block matching algorithms are listed in Table \ref{tab:steg_bma}.

We observe from Table \ref{tab:steg_bma} that all steganalytic features are insensitive to the change of block matching algorithms, and the fluctuations of their accuracy rates are all very small, indicating that the conventional local optimality and our generalized local optimality are robust to the block matching algorithms. The accuracy rates of GLO and its sub-features still generally outperform those of NPE under various QPs and payloads. For each type of features, the accuracy rate generally increases in the order of DIA, HEX, UMH, and ESA. The DIA, HEX and UMH are fast search methods, and their search accuracy increases gradually; ESA is a global search with the highest accuracy. More accurate block matching algorithm leads to more accurate rate-distortion costs, which are conducive to steganalysis.

\subsubsection{Steganalysis on Videos with Different Resolutions}
The videos in the real-world have various resolutions, which affect the texture complexity and thus the rate-distortion cost. To evaluate the steganalytic features on different videos resolutions, we perform steganalysis experiments on DB3 at QP 15 and 25. The videos in DB3 have a resolution of 1920$\times$1080, which is quite different from the CIF videos used in other experiments. The accuracy rates of NPE, GLO-MV, GLO-PMV and GLO for the videos in DB3 are listed in Table \ref{tab:steg_resolution}.

We observe that the all the methods obtain promising performance on DB3, and the accuracy rates are higher than those on DB1. This is because the high-resolution videos have relatively more smooth textures, which are easy for block matching in ME and thus can improve the accuracy of rate-distortion costs for steganalysis.

\subsubsection{Steganalysis for Different Video Coding Standards}
The (generalized) local optimality can be applied to any video coding standards that adopt rate-distortion optimization. To evaluate the steganalytic features on different video coding standards, we generate the video sequences using an open source MPEG-4 \cite{MPEG2001Draft} encoder called XviD \cite{xvid} with Advanced Simple Profle. The videos in previous experiments are compressed with constant quantization parameter (CQP). To evaluate the performance on dynamic quantization parameters, we compress the videos from DB1 with constant bit-rate (CBR), including 500 kb/s and 1000 kb/s. The accuracy rates of NPE, GLO-MV, GLO-PMV and GLO for MPEG-4 compressed videos are reported in Table \ref{tab:steg_mpeg4}.

It can be seen from Table \ref{tab:steg_mpeg4} that all the methods can still achieve satisfactory results under the MPEG-4 standard. The accuracy rates for MPEG-4 are all slightly lower than those for H.264/AVC, and this is caused by the difference of video codecs, in which the ME in MPEG-4 is less accurate than that in H.264/AVC. In addition, the detection performance of GLO and its sub-features is generally better than that of NPE, indicating that the advantages of generalized local optimality can also be maintained in other video coding standards.

\begin{table}[t]
  \newcommand{\tabincell}[2]{\begin{tabular}{@{}#1@{}}#2\end{tabular}}
  \caption{Computational Time (in Second) of Feature Extraction}
  \vspace{-5pt}
  \label{tab:fea_time}
  \centering
  \begin{tabular}{ccccc}
  \toprule
  \multirow{2}{*}[-2pt]{\tabincell{c}{Feature\\set}} & \multicolumn{2}{c}{QP 15} & \multicolumn{2}{c}{QP 25} \\
  \cmidrule(r){2-3} \cmidrule(l){4-5}
  & Time & Ratio & Time & Ratio \\
  \midrule
  AoSO    & 348.65 & 1    & 261.42 & 1    \\
  NPE     & 456.73 & 1.31 & 340.85 & 1.30 \\
  GLO-MV  & 467.19 & 1.34 & 345.09 & 1.32 \\
  GLO-PMV & 379.03 & 1.09 & 279.07 & 1.07 \\
  GLO     & 487.11 & 1.40 & 357.25 & 1.37 \\
  \bottomrule
  \end{tabular}
  \vspace{-5pt}
\end{table}

\vspace{-4pt}
\subsection{Comparison of Computational Complexity} \label{sec:exp_time}

The feature extraction using (generalized) local optimality is equivalent to performing ME at the video decoder, which is a time-consuming process. To evaluate the computational complexity of steganalytic methods, we compare the practical computational time of feature extraction for AoSO, NPE, GLO-MV, GLO-PMV and GLO on DB2, and report the results in Table \ref{tab:fea_time}. The experiments are run on a desktop computer with a 3.2GHz Intel Core i5 CPU and 4 GB RAM.

As shown in Table \ref{tab:fea_time}, the AoSO costs the lowest computational time, since it only calculates the SAD during the feature extraction. The NPE has to calculate both the SAD and SATD based rate-distortion costs, which increase the feature extraction time substantially. Compared with NPE, the computational time of GLO-MV and GLO only slightly increases, and the computational time of GLO-PMV is even less than that of NPE. In the process of calculating rate-distortion costs, the SAD or SATD occupies most of the time, and the bit-rate can be obtained by a look-up table. Although the GLO-MV needs to calculate more SAD and SATD based rate-distortion costs, the SAD and SATD can be reused, and the amount of calculations is the same as that of NPE. For the GLO-PMV, it only needs to calculate the SAD based rate-distortion cost, so its feature extraction time is less than NPE.

\section{Conclusion}
The preservation of local optimality in video coding process and the destruction of local optimality caused by steganographic embedding are a pair of contradictions, which can be used for steganalysis to discover the embedding traces. In this paper, we generalize the convention local optimality in two aspects. Specifically, we first generalize the local optimality from the static estimation using fixed predicted motion vectors (PMVs) to the dynamic estimation using variable PMVs, and then generalize the local optimality from the motion vector (MV) domain to the PMV domain. These two generalizations improve the accuracy of estimation and increase the diversity of local optimality, based on which we design two types of steganalytic features and also propose feature symmetrization rules to reduce feature dimension.

Our proposed features have the same order of dimensions as previous works, but achieve higher detection accuracy against various MV based steganography. Besides, the proposed features are also robust to many real-world environments, such as different video sources, video prediction methods, video codecs and video resolutions. Finally, the proposed features do not improve the computational complexity of feature extraction, which can be used for practical applications.


%

%
%
%
%
%

\ifCLASSOPTIONcaptionsoff
  \newpage
\fi



%
%
%

\bibliographystyle{IEEEtran}
\bibliography{IEEEabrv,glo_1}

%








\end{document}